\documentclass{article}

\usepackage[preprint]{neurips_2025}

\usepackage[utf8]{inputenc} 
\usepackage[T1]{fontenc}    
\usepackage{hyperref}       
\usepackage{url}            
\usepackage{booktabs}       
\usepackage{amsfonts}       
\usepackage{nicefrac}       
\usepackage{microtype}      
\usepackage{xcolor}         
\usepackage{amssymb}
\usepackage{enumitem}
\usepackage{colortbl}
\usepackage{amsthm}
\usepackage{siunitx}
\usepackage{wrapfig}
\usepackage[ruled,vlined]{algorithm2e}
\usepackage{tabularx, multirow, booktabs}
\usepackage{adjustbox}
\usepackage{graphicx}
\usepackage{wrapfig}
\usepackage[most]{tcolorbox}

\usepackage{pifont}
\usepackage{xcolor}

\definecolor{CheckGreen}{HTML}{2E7D32}
\definecolor{CrossRed}{HTML}{C62828}
\definecolor{PartAmber}{HTML}{F39C12}

\title{AutoResearchBench: Benchmarking AI Agents on Complex Scientific Literature Discovery}

\author{
AutoResearchBench Team\thanks{\scriptsize
\textbf{Core Contributors:} Lei Xiong (xiongxiongleilei@ruc.edu.cn)$^*$ , Kun Luo (luokun695@gmail.com)$^*$ ($^*$: equal contribution), Ziyi Xia, Wenbo Zhang, Jin-Ge Yao, Zheng Liu.
\textbf{Participants:} Jingying Shao, Jianlyu Chen, Hongjin Qian, Xi Yang, Qian Yu, Hao Li, Chen Yue, Xiaan Du, Yuyang Wang, Yesheng Liu, Haiyu Xu.
\textbf{Supervision:} Zhicheng Dou (dou@ruc.edu.cn), Zheng Liu (zhengliu1026@gmail.com).
}
}

\begin{document}

\maketitle

\begin{tcolorbox}[
  width=\textwidth,
  colback=black!3,
  colframe=white,
  boxrule=0pt,
  arc=2pt,
  fuzzy shadow={2pt}{-2pt}{0pt}{0.4pt}{black!15},
]
\begin{abstract}
Autonomous scientific research is significantly advanced thanks to the development of AI agents. One key step in this process is finding the right scientific literature, whether to explore existing knowledge for a research problem, or to acquire evidence for verifying assumptions and supporting claims. 
To assess AI agents' capability in driving this process, we present \textbf{AutoResearchBench}, a dedicated benchmark for autonomous scientific literature discovery.
AutoResearchBench consists of two complementary task types: (1) \textbf{Deep Research}, which requires tracking down a specific target paper through a progressive, multi-step probing process, and (2) \textbf{Wide Research}, which requires comprehensively collecting a set of papers satisfying given conditions. 
Compared to previous benchmarks on agentic web browsing, AutoResearchBench is distinguished along three dimensions: it is \underline{research-oriented}, calling for in-depth comprehension of scientific concepts; \underline{literature-focused}, demanding fine-grained utilization of detailed full-text information; and \underline{open-ended}, involving an unknown number of qualified papers and thus requiring deliberate reasoning and search throughout. These properties make AutoResearchBench uniquely suited for evaluating autonomous research capabilities, and extraordinarily challenging. 
Even the most powerful LLMs, despite having largely conquered general agentic web-browsing benchmarks such as BrowseComp, achieve only 9.39\% accuracy on Deep Research and 9.31\% IoU on Wide Research, while many other strong baselines fall below 5\%. We publicly release the dataset, evaluation pipeline, and code at \url{https://github.com/CherYou/AutoResearchBench}.
\end{abstract}
\end{tcolorbox}

\section{Introduction}
\label{sec:introduction} 


The recent rise of LLM-based \emph{AI scientists systems} has made autonomous scientific research a concrete target rather than a distant aspiration~\citep{ai_scientist, agent_laboratory, ai_coscientist, tang2025ai, schmidgall2025agent}.
Across ideation, design, implementation, and experimentation, one capability is repeatedly indispensable: finding the right scientific literature~\citep{seo2025paper2code, xu2026idea2story, li2025deepcode, chen2026toward, weng2025deepscientist}. 
Literature discovery serves both to explore existing knowledge around a problem and to gather evidence for verifying assumptions and supporting claims. 
An autonomous researcher must identify what is already known, which assumptions are supported or contradicted, what methods and evaluation protocols are appropriate, and where the strongest evidence for a claim resides. 
Scientific literature discovery is therefore a core operation for autonomous research~\citep{agent_laboratory, deepxiv}. 


In practice, this is often not a matter of retrieving papers about a topic, but of determining whether any paper in a large corpus satisfies a hidden conjunction of technical constraints. 
The challenge stems from both \emph{complexity} and \emph{open-endedness}. 
On the complexity side, the relevant knowledge is technical, fast-moving, and domain-specific rather than common-sense~\citep{litsearch, rpcbench}. 
Moreover, decisive evidence is rarely apparent in titles, abstracts, or short snippets; instead, it is typically distributed in method details, ablation tables, figure captions, appendices, and citation chains across full paper context.  
On the open-endedness side, researchers frequently search with assumption-based conditions whose satisfiability is itself unknown. 
The agent may need to conclude that no paper qualifies, identify a unique target paper, or exhaustively recover all papers satisfying a multi-part scientific condition. This requires deliberate reasoning about both correctness and completeness~\citep{litsearch, wong2025widesearch}. 

These properties make scientific literature discovery qualitatively different from general web browsing. 
Recent agents have made strong progress on long-horizon web-search benchmarks such as GAIA and BrowseComp~\citep{gaia, browsecomp, BrowseComp-Plus, wu2025webdancerautonomousinformationseeking, li2025websailor, luo2025infoflow, searcho1, searchr1}. 
What remains largely unmeasured is whether an agent can search a large, up-to-date scientific corpus, read full papers in depth, verify fine-grained technical conditions, and decide when its search is complete~\citep{qasper, sage, litqa2, wang2025paperarena, he2025pasa}. 
As a result, current benchmark performance gives only a partial picture of an agent's readiness for autonomous research. 

\begin{figure}[t]
  \centering
  \vspace{-12pt}
  \includegraphics[width=0.9\linewidth]{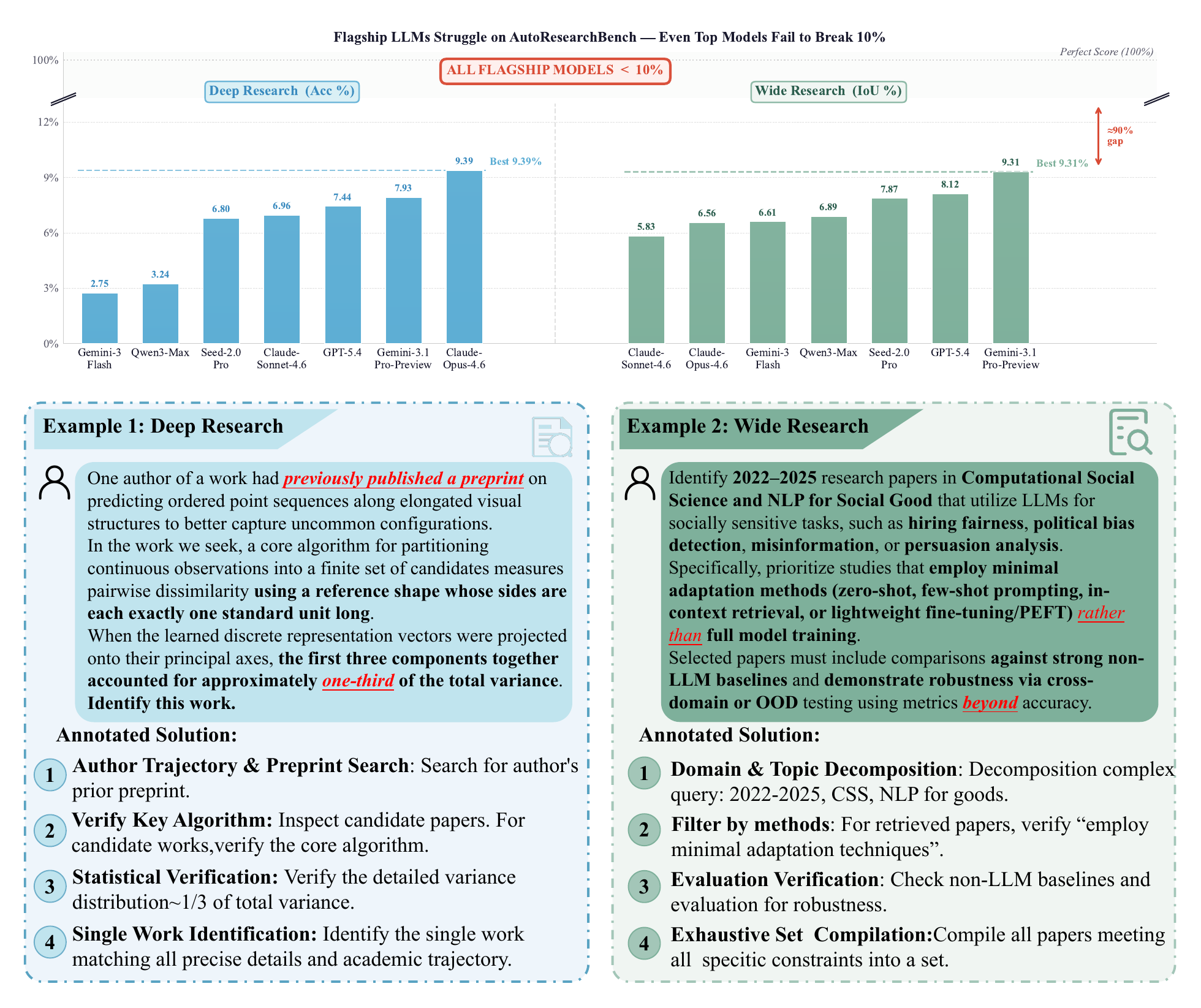}
  \vspace{-4pt}
  \caption{\textbf{All Flagship models Struggle on AutoReasearchBench.} Two representative (Deep Research and Wide Research) instances showing multi-hop trajectory reasoning, fine-grained detail verification, and complex constraint decomposition (e.g., iterative web search and full-text reading) to derive a verifiable unique target paper or an exhaustive literature set.}
  \vspace{-20pt}
  \label{fig:case}
\end{figure}

In this paper, we introduce \textbf{AutoResearchBench}, a benchmark for autonomous scientific literature discovery. 
AutoResearchBench contains 1,000 problems spanning eight core areas of computer science, built through a full-text-first human--machine pipeline and verified by human experts. 
It contains two complementary task types. 
\textbf{Deep Research} requires tracking down a specific target paper through a progressive, multi-step probing process, or concluding that no paper satisfies the condition. 
\textbf{Wide Research} requires comprehensively collecting the set of papers that satisfy a given scientific specification. 
Together, these tasks capture two recurring demands in real research workflows: precise paper identification and comprehensive literature coverage. Figure~\ref{fig:case} shows representative examples of AutoResearchBench.

AutoResearchBench is distinguished by three properties. 
It is \emph{research-oriented}: solving tasks demands in-depth understanding of scientific concepts rather than shallow matching or common-sense reasoning. 
It is \emph{literature-focused}: critical clues come from fine-grained paper content, including tables, figures, and reference lists, rather than only metadata or abstracts. 
And it is \emph{open-ended}: the number of qualifying papers is unknown, sometimes zero, so agents must reason carefully about exploration, verification, abstention, and stopping. 
To support rigorous evaluation, we build a controlled environment over more than three million arXiv papers with up-to-date full-text extraction and search tools~\citep{deepxiv}, enabling agents to operate over realistic scientific evidence rather than a simplified retrieval setup. 



Experiments based on ReAct~\citep{yao2023react} framework show that AutoResearchBench is extraordinarily challenging. Frontier models that perform strongly on general browsing benchmarks achieve only 9.39\% accuracy on Deep Research and 9.31\% IoU on Wide Research, while many strong baselines fall below 5\%. This contrasts sharply with BrowseComp~\citep{browsecomp}, revealing a distinct performance gap in scientific literature search uncaptured by general web benchmarks. Longer trajectories, increased reasoning, and additional tool calls yield limited gains. Dominant errors reflect deeper limitations: weak scientific reasoning, incomplete use of paper-level information, difficulty handling long conjunctive queries, insufficient comprehensiveness in set discovery, and limited reflection during iterative search. These findings suggest scientific literature discovery is not merely a harder instance of web browsing, but a distinct capability frontier for autonomous research agents.

Our contributions are as follows:

(1) We introduce AutoResearchBench, a new benchmark for autonomous scientific literature discovery with two task families, Deep Research and Wide Research, that capture precise paper identification and comprehensive literature collection. 

(2) We construct real and challenging tasks through a full-text-first human--machine pipeline, and provide high-quality gold answers verified by human experts. 

(3) We build an evaluation infrastructure over an up-to-date arXiv corpus, including extracted full-paper content and search tools, for controlled large-scale evaluation. 

(4) We provide a comprehensive empirical study showing that current frontier agents remain far from saturation, identify the main bottlenecks future systems must address, and publicly release the dataset and evaluation pipeline. 

\vspace{-3pt}

\section{AutoResearchBench}
\label{sec:academic_SAC}
We introduce \textbf{AutoResearchBench}, a benchmark for evaluating AI agents on complex scientific literature discovery.
The benchmark comprises two complementary task paradigms, \emph{deep research}, i.e., identifying a uniquely correct paper from non-salient full-text evidence, and \emph{wide research}, i.e., exhaustively covering a technical slice of the literature under precise scientific constraints. 
We next describe the task formulation, construction pipeline, dataset statistics, and evaluation protocol. 

\vspace{-3pt}
\subsection{Task Creation}
AutoResearchBench evaluates whether an agent can autonomously search, browse, and reason over scientific papers to find targets. 
Given a query $q$ and a corpus $\mathcal{D}$, the agent interacts with a search environment and outputs a predicted answer set $\hat{Y}(q)$. Formally, at step $t$ the agent state is
$s_t = (q, h_t, \mathcal{D}_t)$,
where $h_t$ is the interaction history and $\mathcal{D}_t \subseteq \mathcal{D}$ is the set of documents observed so far. The agent conducts reasoning and uses the search tool (DeepXiv search~\citep{deepxiv} or web search) or chooses a terminal answer action, and after a finite trajectory outputs $\hat{Y}(q)$ intended to approximate
$Y^*(q) = \{ d \in \mathcal{D} : d \models q \}$.
Unlike general web QA or context-provided scientific QA, the decisive evidence here often lies in appendices, ablations, proof details, figure captions, or citation contexts rather than titles and abstracts.

We instantiate two complementary task paradigms: 

\textbf{Academic Deep Research.} Deep Research systematically evaluates the precise identification capability of intelligent agents. 
Here $|Y^*(q)| \in \{0,1\}$: the agent must either identify the unique paper satisfying the query or rigorously establish that no such paper exists. 
Queries are carefully constructed from tightly coupled and deliberately obfuscated constraints derived from full-text information and citation relationships. While each individual clue is weak in isolation, the conjunction of these clues is highly discriminative. Consequently, successful completion of the task requires persistent full-text browsing, iterative hypothesis refinement, and multi-hop scientific reasoning across documents.

\textbf{Academic Wide Research.} Wide Research evaluates exhaustive coverage. It asks the agent to retrieve the complete set of papers satisfying a scientific query. 
The challenge is not to find one representative paper, but to trace the boundary of the target concept through a large corpus while preserving precision. The task therefore stresses both recall-oriented exploration and evidence-based filtering.
Together, the two paradigms capture the two research workflows emphasized in the introduction: precise paper identification and comprehensive literature coverage. 

\subsection{Construction Pipeline}
\begin{figure}[ht]
    \centering
    \includegraphics[width=0.9\linewidth]{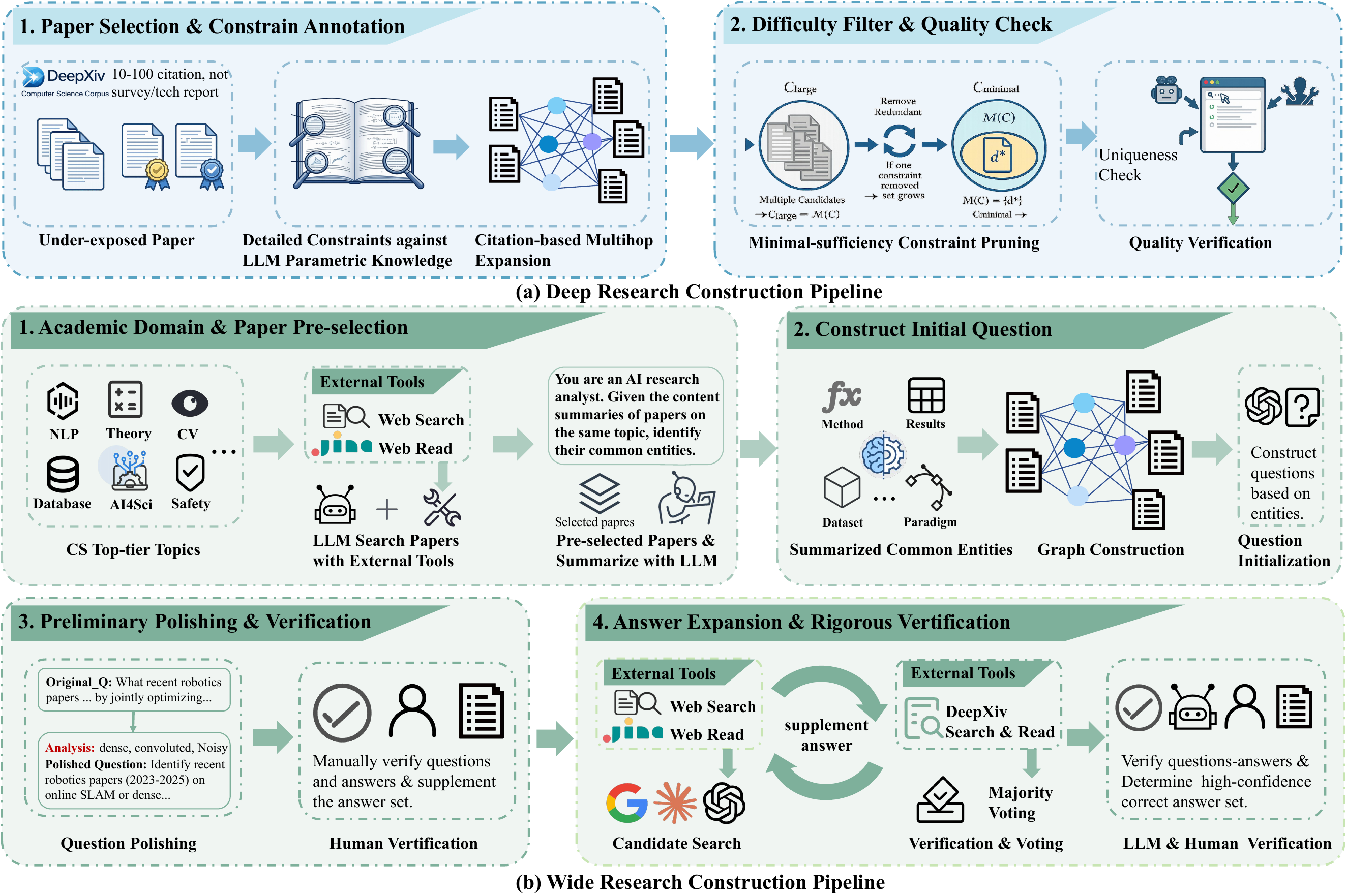}
    \caption{Overview of the benchmark construction pipeline. The construction pipeline comprises 2 stages: (a) The Deep Research construction pipeline;(b) The Wide Research construction pipeline.}
    \label{fig:pipeline}
\end{figure}

To construct challenging and realistic academic search tasks, we design a multi-stage pipeline with human involvement at critical steps, as illustrated in Figure~\ref{fig:pipeline}. The process builds on a curated scientific corpus and combines model-assisted generation with manual verification. Models equipped with rich knowledge can improve the efficiency and difficulty of task construction, while verification by human experts ensures that the questions conform to real-world scientific preference.

The design is tailored to the demands of scientific literature discovery, including long-horizon exploration, full-text inspection, and scientific reasoning. 

We build the benchmark on a controlled corpus of over three million papers hosted by the DeepXiv~\citep{deepxiv}, an open-access academic platform. This corpus offers agentic search tool, structured metadata and full-text access, making it well suited for search agents that require long-horizon exploration, full paper browsing and scientific reasoning.

\subsubsection{Academic Deep Research Task Construction Pipeline}
\label{sec:pipeline_deep_search}

Academic Deep research tasks are built via a full-text-first pipeline to avoid shallow metadata matching. For each target paper $d^* \in \mathcal{D}$, we extract and iteratively refine natural-language constraints $C = \{c_i\}_{i=1}^m$ from full-text evidence, narrowing the feasible set
$\mathcal{M}(C) = \{d \in \mathcal{D} : d \models C\}$,
where $d \models C$ indicates paper $d$ satisfies all constraints in $C$, into the singleton $\{d^*\}$.
This design follows the \emph{minimal sufficiency} principle: constraints uniquely locate the target, and removing any yields alternative papers. To construct no-answer instances, annotators perturb one core constraint (e.g., method, time, or author) so no valid paper exists in the corpus. These negative cases test whether agents can abstain from plausible but invalid matches. As shown in Figure~\ref{fig:pipeline} (a), the whole construction pipeline includes four stages.

\textbf{(1) Target Paper Selection}. We sample technically substantive computer science papers that are unlikely to be recovered by memorization alone, preferentially focusing on under-exposed yet high-quality works (typically with 10--100 citations). We exclude surveys and broad technical reports, whose scope weakens paper-specific identifiability.

\textbf{(2) Full-Text Constraint Mining and Citation-based Multi-hop Expansion}. Annotators carefully read the full paper, including supplementary appendices when necessary, and extract verifiable clues from subtle secondary methodological choices, rigorous proof or derivation details, fine-grained local empirical observations, and author--organization affiliation relations.

We explicitly avoid headline cues such as dataset names, main claims, and core contributions. Citation relations are promoted into explicit subproblems by extracting supporting clues from cited papers when appropriate. This transforms simple paper lookup into multi-hop academic search.

\textbf{(3) Constraint Fuzzification and Pruning}. The raw clue pool is rewritten to suppress lexical shortcuts. We apply fuzzification at two levels: topic-level fuzzification, which avoids explicitly revealing the research area, and detail-level fuzzification, which paraphrases local evidence to reduce keyword hits. 
We then perform iterative pruning by testing the evolving query against the corpus, removing redundant constraints and keeping only those that materially contribute to uniquely identifying the target paper. This step operationalizes the \textbf{minimal-sufficiency objective}.

\textbf{(4) Verification}. Each instance is repeatedly solved by frontier LLM agents and human annotators in the same DeepXiv environment used at evaluation time. We retain an instance only if (i) the gold answer is supported by explicit textual evidence for answerable cases, or no valid paper can be found after corpus-level checking for no-answer cases, (ii) plausible alternatives can be systematically ruled out, and (iii) no shallow reformulation makes the task trivial. Disagreements trigger another round of reading, query revision, and corpus-level checking.

\subsubsection{Academic Wide Research Task Construction Pipeline}
\label{sec:pipeline_wide_search}
Unlike deep research's unique identification, Academic wide research requires searching an exhaustive set of relevant documents. Formally, for a query $q$, the objective is to approximate the complete valid set $Y^*(q) = \{ d \in \mathcal{D} \mid d \models q \}$, where $\models$ denotes that document $d$ satisfies all implicit constraints in $q$. To maximize coverage while maintaining strict semantic consistency, we design a four-stage construction pipeline grounded in academic entity graphs (Figure~\ref{fig:pipeline} (b)).

\textbf{(1) Domain-Specific Candidate Sourcing.} We define high-level research themes across core CS domains. Using external search tools, we retrieve preliminary candidate pools, which are subsequently filtered and summarized by an LLM to ensure topical cohesion and representativeness.

\textbf{(2) Structural Abstraction and Query Formulation.} We extract shared multidimensional attributes (e.g., method, datasets, results) from the candidates to construct an entity graph. This structural prior is then translated into an initial query encoding a strict conjunction of these shared constraints.

\textbf{(3) Query Refinement and Initial Verification.} The query is rewritten into a natural scientific intent while strictly preserving its logical constraints. Human annotators verify the alignment between the query and the candidate set, manually augmenting it with missing valid papers to establish a reliable initial approximation of $Y^*(q)$.

\textbf{(4) Iterative Expansion and Rigorous Auditing.} To approach absolute completeness, we iteratively expand the set via search tools. Newly retrieved documents undergo full-text analysis and are evaluated by an ensemble of three advanced LLMs; a paper is admitted only upon unanimous consensus. This expansion halts when no new valid candidates emerge. Finally, human experts conduct a meticulous audit of the final sets, purging any documents that marginally violate the constraint conjunction. Details are in Appendix~\ref{appendix:detailed verification}.

\subsection{Benchmark Statistics}

\begin{figure}[t]
    \centering
    \includegraphics[width=0.9\linewidth]{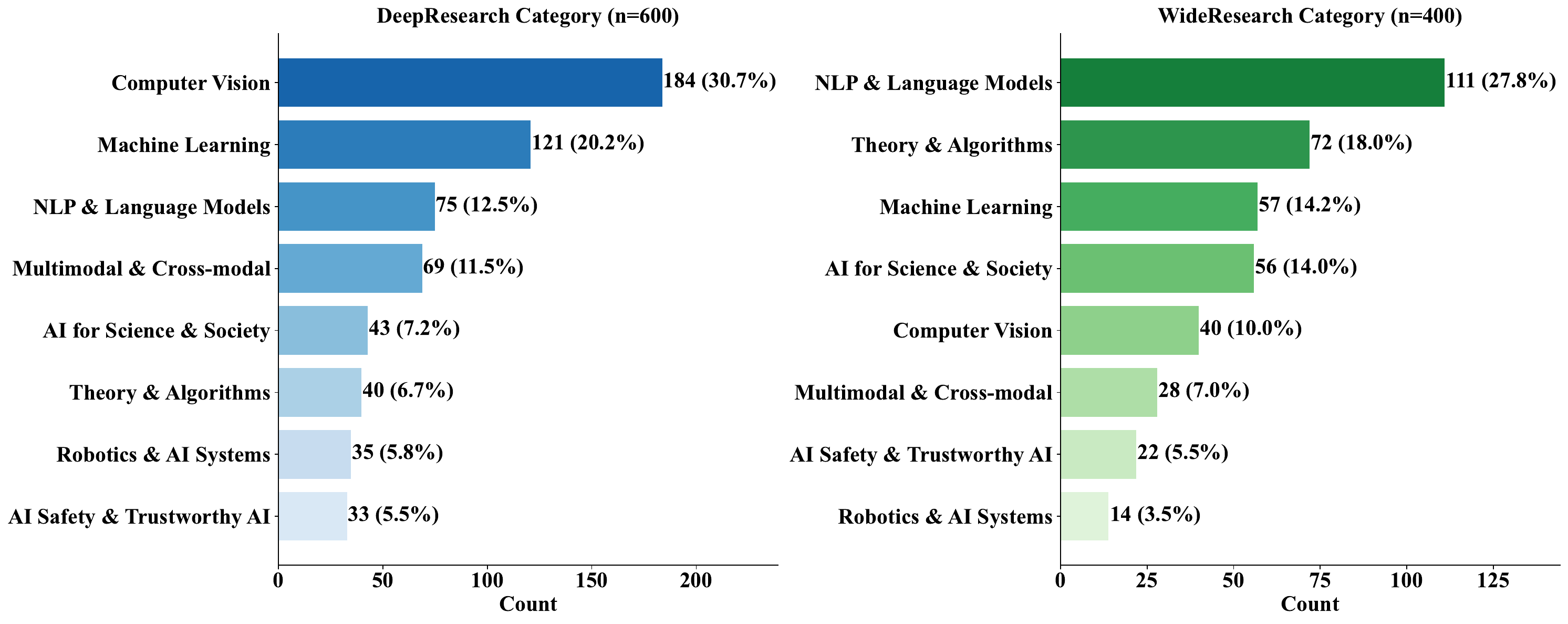}
    \caption{Category distribution of two tasks across major computer science domains.}
    \label{fig:category_distribution}
\end{figure}

The finalized benchmark comprises 1,000 high-quality queries (600 Deep, 400 Wide) spanning eight core computer science domains, as illustrated in Figure~\ref{fig:category_distribution}. As detailed in Table~\ref{tab:dataset_stats}, these two task paradigms evaluate fundamentally different search capabilities via distinct answer space topologies. Deep Research primarily assesses precise constraint satisfaction: 90\% of its queries target a single valid paper, while 10\% are intentionally unsatisfiable (zero answers) to test the agent's capacity for recognizing infeasible constraints. Wide Research focuses on measuring systematic coverage, encompassing 3,692 relevant papers with an average of 9.23 valid answers per query. This task requires a robust balance between broad exploration and strict filtering: on average, each query initially retrieves 33 candidate papers, which must undergo rigorous multi-stage iterative verification and filtering to yield the final exact answer sets. Given the substantial annotation costs of this construction process, detailed statistics and cost analyses are provided in Appendix~\ref{appendix:detailed statistic}.

\subsection{Evaluation Metrics}

To ensure reproducible evaluation, we deploy a standardized ReAct-based agent~\cite{yao2023react} that interacts with a curated arXiv corpus via a unified \texttt{DeepXiv search} tool, an efficient and high-performance agentic literature search tool built upon the DeepXiv platform.. Further implementation details and comparisons with open-web settings are deferred to Section~\ref{subsec:tool_comparison}. Since our benchmark comprises two distinct task paradigms, we formalize tailored evaluation metrics for each.

\begin{wraptable}{r}{0.6\textwidth}
    \vspace{-10pt}
    \centering
    \small
    \caption{Statistics of query and answer distributions.}
    \label{tab:dataset_stats}
    \begin{tabular}{@{}lcc@{}}
        \toprule
        \textbf{Statistic} & \textbf{Deep Research} & \textbf{Wide Research} \\
        \midrule
        Number of Queries        & 600   & 400   \\
        Total Answers            & 540   & 3692 \\
        Answer Cardinality       & $\{0,1\}$ & $[2,34]$ \\
        Avg.\ Answers per Query  & --    & 9.23 \\
        \bottomrule
    \end{tabular}
\end{wraptable}

\label{subsec:evaluation}

\paragraph{Deep Research (Accuracy).}
Deep research requires precisely isolating a unique target document or correctly determining its absence, such that the ground-truth set satisfies $|Y^*(q)| \in \{0, 1\}$. We adopt \textbf{Accuracy} to evaluate this exact-match objective. Given a predicted set $\hat{Y}(q)$ and the ground-truth $Y^*(q)$, the accuracy over a query set $\mathcal{Q}$ is defined as:
\begin{equation}
\text{Acc} = \frac{1}{|\mathcal{Q}|} \sum_{q \in \mathcal{Q}} \mathbb{I}(\hat{Y}(q) = Y^*(q))
\end{equation}
where $\mathbb{I}(\cdot)$ is the indicator function. This strict metric penalizes any partial correctness, directly reflecting the agent's ability to resolve tightly coupled constraints.

\paragraph{Wide Research (IoU).}
Wide research is formulated as a constrained set completion task, where $Y^*(q)$ typically contains multiple valid candidates. To jointly assess precision and coverage (recall) without imposing an artificial ranking order, we utilize \textbf{Intersection over Union (IoU)}:
\begin{equation}
    \text{IoU} = \frac{1}{|\mathcal{Q}|} \sum_{q \in \mathcal{Q}} \frac{|\hat{Y}(q) \cap Y^*(q)|}{|\hat{Y}(q) \cup Y^*(q)|}
\end{equation}

Unlike traditional top-$k$ or ranking-based search metrics, IoU evaluates the predicted set holistically. It rigorously requires the agent to systematically enumerate all valid candidates while strictly avoiding out-of-scope inclusions, effectively measuring large-scale exploratory reasoning capabilities.

\section{Experiments}
\label{sec:experiments}
\subsection{Experiment Setup}

To evaluate the capability of current systems on academic deep research and wide research, we consider a set of both open-source and proprietary models, covering different scales and settings.

\paragraph{Models.}
Our evaluation covers multiple open-source models, including Qwen3.5-35B-A3B~\cite{qwen3.5}, Qwen3.5-122B-A10B~\cite{qwen3.5}, Qwen3.5-397B-A17B~\cite{qwen3.5}, Deepseek-V3.2~\cite{deepseekai2025deepseekv32}, MiniMax-M2.5~\cite{minimax2026m25} and Kimi-K2.5~\cite{kimiteam2026kimik25visualagentic}, alongside a collection of proprietary models such as Qwen3-max~\cite{yang2025qwen3}, Seed 2.0 pro~\cite{bytedance2026seed2}, Gemini-3-flash~\cite{deepmind2026gemini3flash}, Gemini-3.1-pro~\cite{deepmind2026geminipro}, GPT-5.4~\cite{openai2026gpt54}, Claude-Sonnet-4.6~\cite{anthropic2026sonnet46}, and Claude-Opus-4.6~\cite{anthropic2026opus46}. These candidates correspond to mainstream state-of-the-art large language models with differentiated proficiency in logical reasoning and tool utilization. Furthermore, we conduct quantitative assessments on several end-to-end research systems, namely Alphaxiv~\cite{alphaxiv_website}, GPT DeepResearch~\cite{deepresearch}, and AI Studio Gemini-3.1-pro~\cite{google_ai_studio}, which combine information search and logical reasoning within a unified architectural design.

\paragraph{Implementation.}
We run open-source models via SGLang under a unified setup, while accessing proprietary models through official APIs. All models operate within the same ReAct-based~\cite{yao2023react} agent framework and interact with the \texttt{DeepXiv search} tool from Section~\ref{subsec:evaluation}, ensuring performance differences mainly reflect model capability. For costlier, less accessible end-to-end systems, we evaluate a random sample of 50 queries. Experiments follow default settings, with details in Appendix~\ref{appendix:detailed experiments}.

\begin{figure}[ht]
    \centering
    \includegraphics[width=0.9\linewidth]{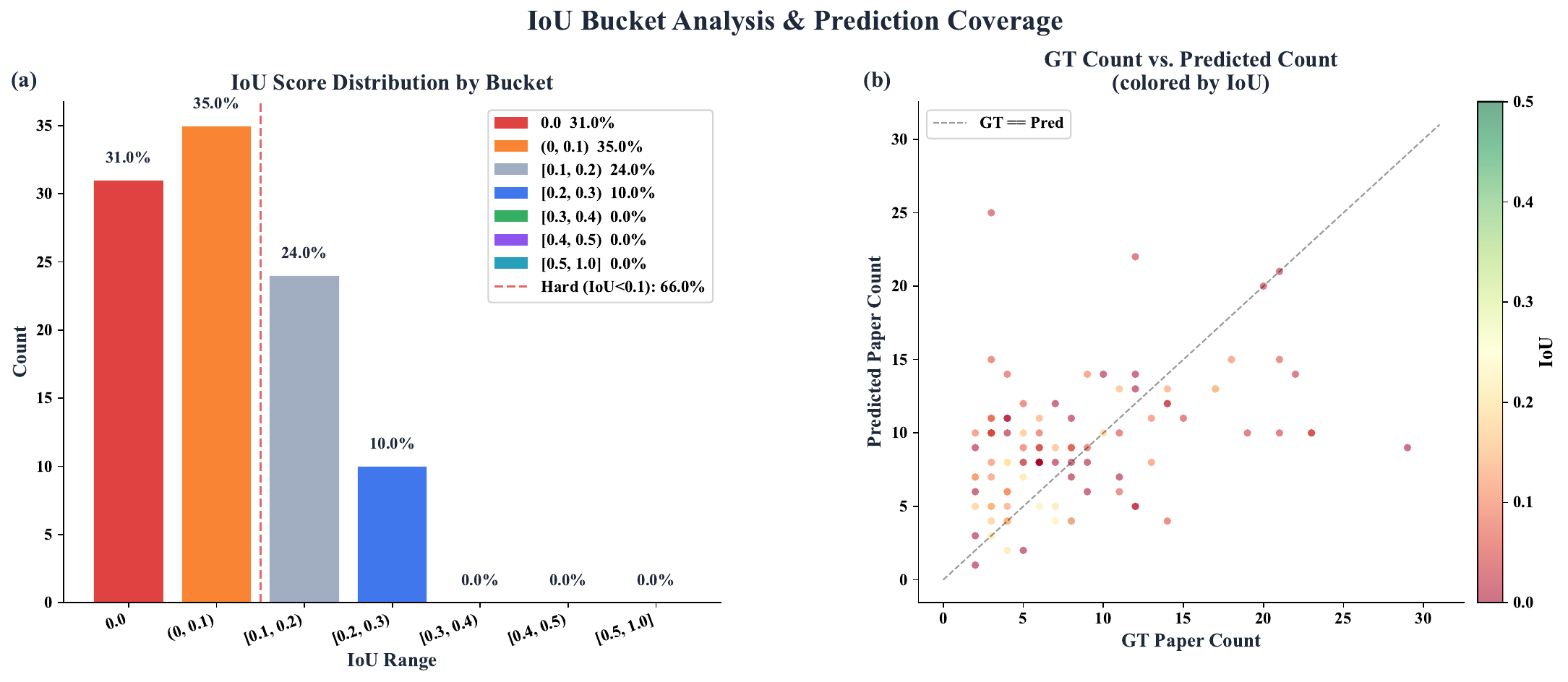}
    \caption{\textbf{Wide Search IoU bucket analysis and prediction coverage of Gemini-3.1-pro (100 cases).} (a) Distribution of IoU scores across different performance. (b) Scatter plot comparing the number of ground truth papers versus predicted papers.}
    \vspace{-10pt}
    \label{fig:bucket_analysis}
\end{figure}

\subsection{Main Results}
\definecolor{DeepColor}{HTML}{C0E4EE}
\definecolor{WideColor}{HTML}{7EB09B}

\begin{table*}[ht]
\centering
\small
\caption{Main experimental results on AutoResearchBench. Models and systems are evaluated using the deepxiv search tool (except End-to-End system). We report Accuracy (\%) for Deep Research and Intersection over Union (IoU) for Wide Research. Darker background colors indicate better performance. The best results are in \textbf{bold}, and the second-best results are \underline{underlined}.}
\label{tab:main_results_grouped}
\resizebox{0.90\textwidth}{!}{%
\begin{tabular}{l ccc c ccc}
\toprule
\multirow{2}{*}{\textbf{Model / System}} & \multicolumn{3}{c}{\textbf{Deep Research (Accuracy)}} & & \multicolumn{3}{c}{\textbf{Wide Research (IoU)}} \\
\cmidrule(lr){2-4} \cmidrule(lr){6-8}
& \textbf{Acc (\%)} & \textbf{Time (s)} & \textbf{Turns} & 
& \textbf{IoU (\%)} & \textbf{Time (s)} & \textbf{Turns} \\
\midrule

\multicolumn{8}{l}{\cellcolor{gray!10}\textit{\textbf{Open-Source Models}}} \\
\midrule
Qwen3.5-35B-A3B & \cellcolor{DeepColor!21}1.94 & 371.0 & 28.9 & & \cellcolor{WideColor!29}2.71 & 274.7 & 12.83 \\
Qwen3.5-122B-A10B & \cellcolor{DeepColor!41}3.88 & 625.0 & 26.2 & & \cellcolor{WideColor!30}2.76 & 594.9 & 5.39 \\
Qwen3.5-397B-A17B & \cellcolor{DeepColor!74}6.97 & 3,230.2 & 27.4 & & \cellcolor{WideColor!41}3.83 & 1,027.3 & 7.11 \\
Deepseek-V3.2 & \cellcolor{DeepColor!45}4.21 & 405.7 & 28.8 & & \cellcolor{WideColor!83}7.70 & 560.5 & 6.25 \\
MiniMax-M2.5 & \cellcolor{DeepColor!31}2.91 & 150.4 & 28.4 & & \cellcolor{WideColor!25}1.39 & 876.9 & 13.09 \\
Kimi-K2.5 & \cellcolor{DeepColor!50}4.69 & 171.9 & 27.0 & & \cellcolor{WideColor!67}6.23 & 353.9 & 8.35 \\
\midrule

\multicolumn{8}{l}{\cellcolor{gray!10}\textit{\textbf{Closed-Source Models}}} \\
\midrule
Qwen3-Max & \cellcolor{DeepColor!35}3.24 & 166.0 & 6.1 & & \cellcolor{WideColor!74}6.89 & 181.9 & 4.20 \\
Seed-2.0-Pro & \cellcolor{DeepColor!72}6.80 & 310.1 & 22.9 & & \cellcolor{WideColor!85}7.87 & 220.6 & 4.15 \\
Gemini-3-Flash & \cellcolor{DeepColor!29}2.75 & 236.9 & 15.9 & & \cellcolor{WideColor!71}6.61 & 64.8 & 4.69 \\
Gemini-3.1-Pro-Preview & \cellcolor{DeepColor!84}\underline{7.93} & 1,221.4 & 24.4 & & \cellcolor{WideColor!100}\textbf{9.31} & 235.3 & 4.55 \\
GPT-5.4 & \cellcolor{DeepColor!79}7.44 & 72.5 & 6.1 & & \cellcolor{WideColor!87}\underline{8.12} & 115.98 & 3.69 \\
Claude-Sonnet-4.6 & \cellcolor{DeepColor!74}6.96 & 588.4 & 27.5 & & \cellcolor{WideColor!63}5.83 & 2,057.0 & 19.0 \\
Claude-Opus-4.6 & \cellcolor{DeepColor!100}\textbf{9.39} & 642.4 & 28.1 & & \cellcolor{WideColor!70}6.56 & 1,032.9 & 27.11 \\
\midrule

\multicolumn{8}{l}{\cellcolor{gray!10}\textit{\textbf{End-to-End Systems}}} \\
\midrule
Alphaxiv & 0/50 & 65.2 & - & & 4.31 & 102.6 & - \\
GPT Deep Research & 11/50 & 1074.1 & - & & 4.06 & 981.4 & - \\
AI-studio Gemini-3.1-Pro & 7/50 & 220.8 & - & & 4.84 & 118.5 & - \\

\bottomrule
\end{tabular}%
}
\vspace{-10pt}
\end{table*}
Table~\ref{tab:main_results_grouped} presents the primary experimental results on the AutoResearchBench.
The findings demonstrate that despite recent advancements in foundation models and agents, especially in web agentic discovery tasks, agentic scientific literature search remains largely unsolved.

\textbf{1. Overall performance remains far below general web-browsing benchmarks.} AutoResearchBench presents a severe challenge. Top-performing systems achieve less than 10\% in both metrics: Claude-Opus-4.6 peaks at 9.39\% accuracy in Deep Research, and Gemini-3.1-Pro-Preview reaches only 9.31\% IoU in Wide Research. The majority of evaluated models, including open-source systems, score below 5\%. This contrasts starkly with performance on general agent benchmarks such as BrowseComp, where scores can exceed 80\%, indicating that models optimized for general web tasks exhibit a substantial performance discrepancy when confronted with the high uncertainty and complex assumptions inherent in realistic scientific search scenarios.

\textbf{2. Increased interaction consumption does not translate to improved outcomes.} Longer trajectories or increased tool calls do not guarantee better performance. For instance, GPT-5.4 achieves a competitive 7.44\% accuracy in Deep Research with only 6.1 turns, whereas Deepseek-V3.2 and Kimi-K2.5 take much longer (28.8 and 27.0 turns) yet yield significantly lower accuracies (4.2\% and 4.69\%). Instead of reducing uncertainty, extended reasoning frequently degrades into illogical continuations. Agents repeatedly examine similar papers, issue redundant queries, or persist in invalid reasoning without acquiring new discriminative evidence. This suggests that \textbf{effective information utilization} is far more critical than expanding the search budget.

\textbf{3. Scientific reasoning over complex constraints constitutes the primary bottleneck.} When crucial evidence is obfuscated or embedded deep in full texts, current agents lack the rigorous, long-horizon reasoning required to isolate a unique target from a massive corpus. Even Claude-Opus-4.6, averaging 28.1 turns per query in Deep Research, fails in over 90\% of cases. In both scenarios, failure rarely stems from omitting relevant literature. Instead, agents typically identify plausible candidates but struggle to precisely verify constraints, eliminate boundary cases, and integrate fragmented evidence to reach logical conclusions.

\textbf{4. Agents lack comprehensiveness and effective iterative reflection.} Unlike general web search, scientific investigation typically requires long, natural-language queries. Applying short-query preferences optimized for web tasks to scientific scenarios causes significant performance degradation. In Wide Research, comprehensive retrieval further demands systematic management of the hypothesis space and result-set completeness. Agents perform poorly when defining the boundaries of technical concepts: Claude-Opus-4.6 expands aggressively over 27.11 turns but fails to filter papers violating strict constraints, yielding a low IoU of 6.56\%. Conversely, models like Seed-2.0-Pro and GPT-5.4 terminate searches prematurely after only 4.15 and 3.69 turns respectively, failing to retrieve the complete set of valid literature. A comprehensive error analysis is deferred to Appendix~\ref{appendix:detailed error analysis}. 

\subsection{Analysis of Different Search Tools}
\label{subsec:tool_comparison}
\begin{table*}[t]
\centering
\vspace{-8pt}
\caption{Analysis of search tools. We compare each model under \textsc{WebSearch} and \textsc{DeepXiv} using the same ReAct agent. We report Accuracy (\%) for Deep Search and IoU (\%) for Wide Search. The better performance within each model-task pair is highlighted in \textbf{bold}.}
\label{tab:search_tools}
\resizebox{\textwidth}{!}{%
\begin{tabular}{l c cccc c cccc}
\toprule
\multirow{2}{*}{\textbf{Model}} & \multirow{2}{*}{\textbf{Tool}} & \multicolumn{4}{c}{\textbf{Deep Search}} & & \multicolumn{4}{c}{\textbf{Wide Search}} \\
\cmidrule(lr){3-6} \cmidrule(lr){8-11}
& & \textbf{Acc (\%)} & \textbf{Tokens} & \textbf{Turns} & \textbf{Calls} &&
\textbf{IoU (\%)} & \textbf{Tokens} & \textbf{Turns} & \textbf{Calls} \\
\midrule
\multirow{2}{*}{Gemini-3-flash}
& \textsc{WebSearch} & 2.01          & 9,574  & 14.30 & 12.90 && 3.99          & 15,517 & 5.15 & 4.08 \\
& \textsc{DeepXiv}   & \textbf{2.75} & 14,052 & 15.90 & 14.90 && \textbf{6.61} & 15,632 & 4.69 & 3.37 \\
\midrule
\multirow{2}{*}{Gemini-3.1-pro}
& \textsc{WebSearch} & 6.82          & 6,744  & 28.50 & 26.50 && 7.37          & 15,466 & 4.08 & 2.92 \\
& \textsc{DeepXiv}   & \textbf{7.93} & 12,206 & 24.40 & 22.80 && \textbf{9.31} & 15,045 & 4.55 & 3.49 \\
\midrule
\multirow{2}{*}{Seed-2.0-pro}
& \textsc{WebSearch} & 3.96          & 14,915 & 22.80 & 21.80 && 4.18          & 17,346 & 5.92 & 4.92 \\
& \textsc{DeepXiv}   & \textbf{6.80} & 14,444 & 22.90 & 21.80 && \textbf{7.87} & 14,537 & 4.15 & 3.13 \\
\midrule
\multirow{2}{*}{Deepseek-V3.2}
& \textsc{WebSearch} & 3.09          & 14,959 & 28.60 & 25.60 && 4.78          & 30,204 & 6.77 & 5.77 \\
& \textsc{DeepXiv}   & \textbf{4.21} & 21,575 & 28.80 & 25.80 && \textbf{7.70} & 25,038 & 6.25 & 5.00 \\
\bottomrule
\vspace{-8pt}
\end{tabular}%
}
\end{table*}
To study the effect of different search tools, we keep the agent framework and models then replace \texttt{DeepXiv search} tool backend with an open-web tool search backend (jina search tool~\footnote{https://s.jina.ai/}). Results are reported in Table~\ref{tab:search_tools}. 
Overall, \texttt{DeepXiv search} tool is more reliable, especially for academic deep search tasks. Averaged over the four matched models, Deep Search accuracy drops from 5.42\% with \texttt{DeepXiv search} to 3.97\% with open-web search. 
This behavior is expected: the academic deep search queries are constructed from paper-internal evidence that is often absent from titles, abstracts, and web-visible summaries, making the target paper harder to hit. 
\texttt{DeepXiv search} tool with full-text-accessible index is substantially better aligned with the task. 
These results suggest that open-web search is noisier, more fragmented, and harder to verify against conjunctions of subtle scientific constraints comparing \texttt{DeepXiv search}. 

\subsection{Analysis of Thinking Modes}
\label{subsec:thinking_modes}
\begin{table*}[t]
\centering
\vspace{-8pt}
\caption{Analysis of thinking modes. We compare each model under \textsc{Think} and \textsc{NoThink} modes using the same ReAct agent and the \texttt{deepxiv search} tool. We report Accuracy (\%) for Deep Search and IoU (\%) for Wide Search. The maximum turn limit is fixed at 30 for all agents. The better performance within each model-task pair is highlighted in \textbf{bold}.}
\label{tab:thinking_modes}
\resizebox{\textwidth}{!}{%
\begin{tabular}{l c cccc c cccc}
\toprule
\multirow{2}{*}{\textbf{Model}} & \multirow{2}{*}{\textbf{Mode}} & \multicolumn{4}{c}{\textbf{Deep Search}} & & \multicolumn{4}{c}{\textbf{Wide Search}} \\
\cmidrule(lr){3-6} \cmidrule(lr){8-11}
& & \textbf{Acc (\%)} & \textbf{Time (s)} & \textbf{Turns} & \textbf{Calls} &&
\textbf{IoU (\%)} & \textbf{Time (s)} & \textbf{Turns} & \textbf{Calls} \\
\midrule
\multirow{2}{*}{Gemini-3-flash}
& \textsc{Think}    & 1.83          & 433.6  & 17.70 & 16.50 && 2.53          & 225.4 & 3.30  & 2.30 \\
& \textsc{NoThink}  & \textbf{2.75} & 236.9  & 15.90 & 14.90 && \textbf{6.61} & 64.7  & 3.12  & 2.12 \\
\midrule
\multirow{2}{*}{Qwen3-max}
& \textsc{Think}    & 2.33          & 170.9  & 6.36  & 4.90  && 4.18          & 217.6 & 12.23 & 2.85 \\
& \textsc{NoThink}  & \textbf{3.24} & 166.0  & 6.10  & 5.10  && \textbf{6.89} & 183.4 & 4.14  & 2.79 \\
\midrule
\multirow{2}{*}{Deepseek-V3.2}
& \textsc{Think}    & \textbf{5.67} & 583.7  & 27.90 & 24.00 && 4.28          & 511.5 & 8.35  & 5.80 \\
& \textsc{NoThink}  & 4.21          & 405.7  & 28.80 & 25.80 && \textbf{5.96} & 206.7 & 6.55  & 4.32 \\
\bottomrule
\vspace{-10pt}
\end{tabular}%
}
\end{table*}
 
We next examine whether explicit reasoning improves agentic academic search. Table~\ref{tab:thinking_modes} compares each model with and without \textsc{Think} under the same ReAct agent and DeepXiv search backend.

We do not observe a consistent benefit from \textsc{Think}. Its impact on Deep Research is unstable across models, and it is generally detrimental for Wide Research. By contrast, \textsc{Think} consistently increases runtime, often leading to more search turns and tool interactions.

These results indicate that additional reasoning does not reliably sharpen the agent's evidence-seeking decisions. Instead, the agent may perform many search and reasoning steps without meaningfully reducing uncertainty, spending extra computation on longer deliberation rather than better evidence collection. This suggests that in literature search, reasoning helps only when directly improving external evidence acquisition; otherwise, it primarily adds latency without performance gains.

\subsection{Analysis of Test-Time-Scaling}
\label{subsec:test_time_scaling}

\begin{figure}[t]
  \centering
  \vspace{-16pt}
  \includegraphics[width=0.90\linewidth]{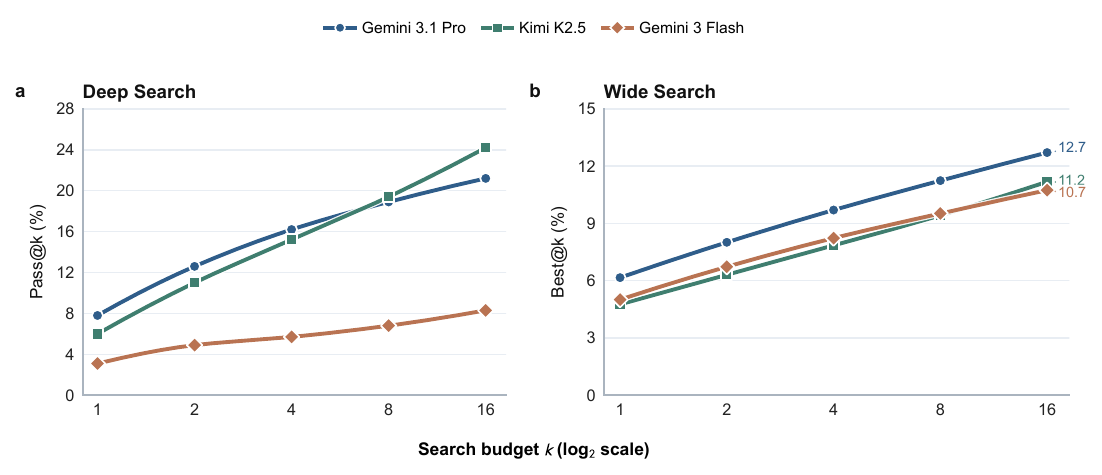}
  \vspace{-10pt}
  \caption{Test time scaling experiment.}
  \label{fig:test_time_scaling}
  \vspace{-10pt}
\end{figure}

We next examine whether test time scaling improves agentic scientific search. Figure~\ref{fig:test_time_scaling} shows results for running each model multiple times under the same agent setup and DeepXiv search backend. For Deep Research, we report pass@$k$; for Wide Research, we report oracle best@$k$ IoU.

Test-time scaling consistently helps, but much more for Deep Research than for Wide Research. This suggests that Deep Research is often limited by brittle trajectory-level decisions: the correct paper is reachable, but a single run may fail to take the right path. By contrast, the smaller gains on Wide Research point to a recall bottleneck, where repeated runs tend to reproduce similar omissions rather than uncover complementary evidence.
We also observe different scaling behaviors across models. kimi-k2.5 benefits more from larger $k$ on Deep Search, while Gemini-3.1-pro remains strongest on Wide Search. Overall, additional inference-time compute is most useful when failures arise from decision instability rather than recall.

\vspace{-8pt}
\section{Related Work}
\vspace{-2pt}
\label{sec:related_work}

\paragraph{Academic Search Agents}
The evolution of autonomous agents has recently shifted towards solving complex, long-horizon information search tasks, giving rise to deep search and deep research paradigms. Agents have demonstrated remarkable capabilities in iterative planning, multi-step reasoning, and web navigation~\cite{searcho1,searchr1,li2025webthinker,deepresearch,google2024deepresearch,tongyidr}. However, open-domain web searches often struggle with the rigorous demands of scientific research, which requires processing peer-reviewed literature, understanding domain-specific terminology, and maintaining strict factual grounding. To bridge this gap, specialized academic search agents have emerged. Efforts like PaSa~\cite{he2024pasa} and SPAR~\cite{shi2025sparscholarpaperretrieval} tailor the search and reasoning pipeline specifically for scholarly contexts, enabling agents to navigate digital libraries, extract methodological details, and synthesize scientific findings with significantly higher reliability.

\paragraph{Benchmark of Academic Search Agents}
The initial agent benchmarks primarily target general-domain tasks. While GAIA~\cite{gaia} and BrowseComp~\cite{browsecomp} assess broad autonomous capabilities, datasets like WideSearch~\cite{wong2025widesearch} and InfoDeepSeek~\cite{xi2025infodeepseek} isolate specific search patterns. Recent works such as DeepWideSearch~\cite{lan2025deepwidesearchbenchmarkingdepthwidth}, GISA~\cite{GISA}, and Table-as-Search~\cite{lan2026tableassearchformulatelonghorizonagentic} attempt to unify deep and wide paradigms. However, lacking controlled corpus, these open-web datasets are unsuitable for the rigorous, evidence-based evaluation required in scientific research.
Within the academic domain, systematic evaluation remains scarce. Datasets like RealScholarQuery~\cite{he2024pasa} and SPAR~\cite{shi2025sparscholarpaperretrieval} provide domain-specific challenges but are severely limited in scale. Conversely, larger benchmarks like SAGE~\cite{hu2026sagebenchmarkingimprovingretrieval} feature controlled, multi-document corpora but lack interactive environments to evaluate dynamic agents. To bridge this gap, we propose AutoResearchBench. Comprising 1,000 curated instances, AutoResearchBench is the first large-scale benchmark designed to systematically evaluate dynamic agents across both deep and wide search paradigms within a controlled scholarly corpus.

\section{Conclusion}
\label{sec:conclusion}

We introduced the AutoResearchBench to systematically evaluate autonomous agents on rigorous scientific literature investigations. Comprising 1,000 expert-curated queries grounded in a controlled, contamination-resistant corpus of over 3 million full-text papers, AutoResearchBench explicitly tests the intersection of long-horizon document browsing and complex scientific reasoning across Deep and Wide Research paradigms. Our comprehensive evaluation of frontier foundation models and end-to-end systems reveals a severe performance gap in real-world academic tasks: state-of-the-art performance peaks at a mere 9.39\% accuracy for Deep Research and 9.31\% IoU for Wide Research. By providing a clean evaluation ecosystem that exposes the insufficiency of shallow heuristic matching and general web navigation, AutoResearchBench establishes a rigorous diagnostic foundation for developing the next generation of reasoning-driven academic agents. 




\bibliographystyle{unsrt}
\bibliography{custom}
\clearpage

\appendix
\section{Ethic Statement}
This benchmark is constructed from publicly accessible scientific papers in arXiv/DeepXiv and does not involve private personal data or direct interaction with human subjects. Human annotators and model assistance are used only for query construction and verification, with final decisions grounded in document evidence. We view AutoResearchBench as an evaluation resource for improving reliable scientific search rather than replacing expert judgment. At the same time, stronger literature-search agents may enable large-scale evidence aggregation, so their deployment should remain transparent, properly cite sources, and include human oversight in high-stakes scientific or policy settings.

\section{Limitations and Future Works}
\label{appendix:limitations}
AutoResearchBench is intentionally designed as a scientific papers discovery benchmark, but this scope also limits its coverage: it currently focuses on computer science papers in a fixed corpus and mainly evaluates text-based search and reasoning, leaving cross-domain science, multi-modal evidence, and continually evolving literature for future study. In addition, although we adopt rigorous verification, exhaustive answer sets for wide search may still be challenging at the boundary of large corpora. Future work can expand AutoResearchBench to broader disciplines, dynamic corpus, and richer tool environments, while also studying calibration, abstention, and more efficient strategies for long-horizon academic search. 

\section{Detailed Statistic of AutoResearchBench}
\label{appendix:detailed statistic}

In this section, we present additional details regarding the statistical characteristics of our benchmark.

\begin{figure}[ht]
    \centering
    \includegraphics[width=\linewidth]{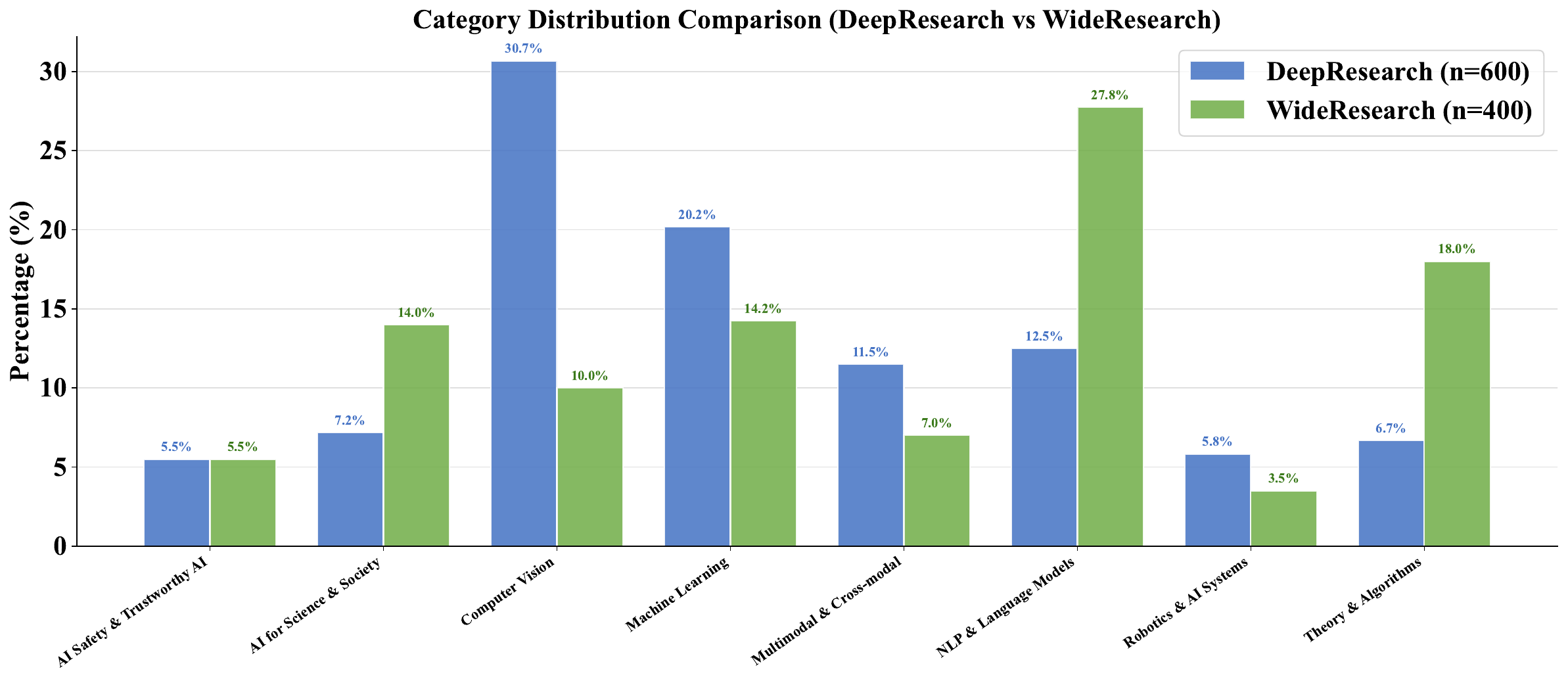}
    \caption{The topic distribution comparison of two tasks.}
    \label{fig:topic_comparison}
\end{figure}

\paragraph{Topic Coverage.}
As shown in Figure.~\ref{fig:topic_comparison}, the topical distributions of \textit{Deep Research} and \textit{Wide Research} differ in proportion yet span all eight major research areas. This demonstrates that both tasks provide broad and representative coverage of key subfields within the computer science domain.

\begin{figure}[ht]
    \centering
    \includegraphics[width=\linewidth]{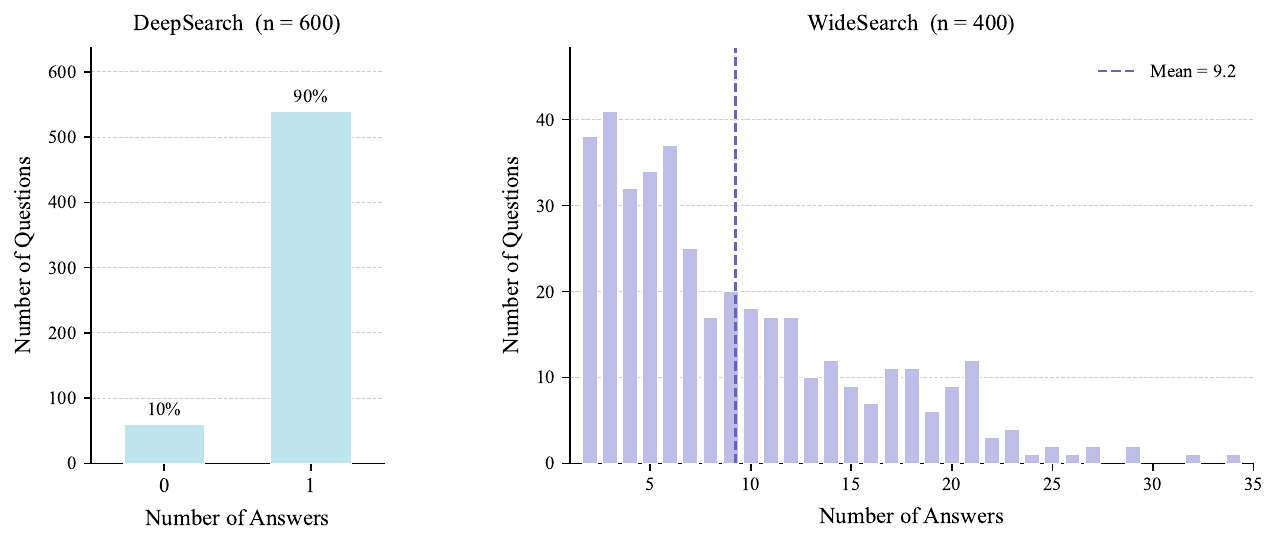}
    \caption{Statistics of the answers per query. (a) Distribution of Deep Search Tasks. (b) Distribution of Wide Search Tasks.}
    \label{fig:answer_distribution}
\end{figure}

\paragraph{Answer Distribution.}
To offer a more intuitive understanding of how answers are distributed across the benchmark, Figure~\ref{fig:answer_distribution} illustrates the answer cardinality patterns for the two task types. For \textit{Deep Research}, we intentionally design approximately 10\% of the queries to have no correct answers within the three-million-paper corpus, enabling us to assess whether models genuinely comprehend query semantics rather than relying on heuristic matching. In contrast, the answer distribution of \textit{Wide Research} exhibits a pronounced long-tail pattern: most queries contain fewer than ten valid answers, while popular research areas may yield more than twenty candidates. This distribution reflects the stringent requirements of Wide Research with respect to satisfying all specified constraints and ensuring completeness of the retrieved answer set.

\paragraph{Cost Statistic.} 
Developing and evaluating AutoResearchBench necessitated an unprecedented scale of computational resources, financial investment, and expert human labor. Unlike standard benchmarks that rely on static metadata or automated backward-construction, our protocol strictly demands full-text comprehension, dynamic tool interactions, and rigorous multi-model consensus. The expenditure is systematically categorized into two primary phases: the Construction and Verification phase, and the Evaluation phase.

The dataset construction pipeline incurred substantial costs due to the deployment of flagship large language models (e.g., Gemini-3.1-pro~\cite{deepmind2026geminipro}, Claude 4.6-Sonnet~\cite{anthropic2026sonnet46}, GPT-5.4~\cite{openai2026gpt54}) for exhaustive candidate generation and zero-noise filtering. For Wide Research tasks, the multi-model voting mechanism required processing dozens of full-length academic papers per query, leading to massive token consumption. Concurrently, Deep Research tasks demanded prolonged autonomous exploration trajectories to locate singular verifiable targets. This computational overhead was heavily supplemented by costly human-in-the-loop expert verification to guarantee absolute factual correctness and eliminate data contamination. Table~\ref{tab:cost_construction} details the quantitative resource breakdown for the dataset.

\begin{table}[htbp]
\centering
\caption{Resource Expenditure for Dataset Construction and Verification. Token counts and API costs reflect the aggregate consumption across multiple flagship models utilized during the multi-stage pipeline. Human effort represents hours spent by domain experts on query polishing and golden document validation.}
\label{tab:cost_construction}
\resizebox{\textwidth}{!}{%
\begin{tabular}{lccccc}
\toprule
\textbf{Task Paradigm} & \textbf{Instances} & \textbf{Avg. Tokens / Query} & \textbf{Avg. Tool Calls / Query} & \textbf{Est. API Cost (\$)} & \textbf{Expert Labor (Hours)} \\
\midrule
Deep Research & 600 & $\sim$170,000 & 28.6 & $\sim$3,500 & $\sim$300 \\
Wide Research & 400 & $\sim$260,000 & 15.2 & $\sim$3,200 & $\sim$280 \\
\midrule
\textbf{Total} & 1000 & - & - & $\mathbf{\sim 6,700}$ & \textbf{$\sim$580} \\
\bottomrule
\end{tabular}%
}
\end{table}

The evaluation phase introduced an additional layer of exponential resource consumption. To establish comprehensive baselines, we assessed over 10 state-of-the-art models and research agents. Assessing a single model across the entire benchmark requires the agent to autonomously navigate the DeepXiv corpus, interacting with external search and reading tools dynamically. The open-ended nature of Wide Research boundaries and the multi-hop reasoning requirements of Deep Research inherently trigger extended observation loops. Consequently, executing a single full-scale evaluation run for one model generates millions of input and output tokens. The detailed average operational costs incurred during the evaluation phase are summarized in Table~\ref{tab:cost_evaluation}.

\begin{table}[htbp]
\centering
\caption{Average Computational Cost per Model Evaluation Run. The statistics represent the mean expenditure required to benchmark a single flagship LLM across the entire dataset.}
\label{tab:cost_evaluation}
\resizebox{\textwidth}{!}{%
\begin{tabular}{lcccc}
\toprule
\textbf{Task Paradigm} & \textbf{Evaluated Models} & \textbf{Avg. Tokens / Query} & \textbf{Avg. Steps / Query} & \textbf{Avg. API Cost / Model Run (\$)} \\
\midrule
Deep Search & 10 & $\sim$21,068 & 22.90 & $\sim$11 \\
Wide Search & 10 & $\sim$32,564 & 9.25 & $\sim$16 \\
\midrule
\textbf{Aggregate (Per Model)} & - & - & - & $\mathbf{\sim 27}$ \\
\bottomrule
\end{tabular}%
}
\end{table}

Cumulatively, the creation and benchmarking of AutoResearchBench represent an order-of-magnitude increase in evaluation rigor, translating directly into significant empirical validation costs. These statistics underscore the inherent difficulty of the benchmark and the necessity of such rigorous, resource-intensive environments to accurately measure true agentic capabilities in authentic scientific research scenarios.

\section{Detailed Verification}
\label{appendix:detailed verification}

\subsection{Deep Research Verification Pipeline}

\begin{figure}[ht]
    \centering
    \includegraphics[width=0.85\linewidth]{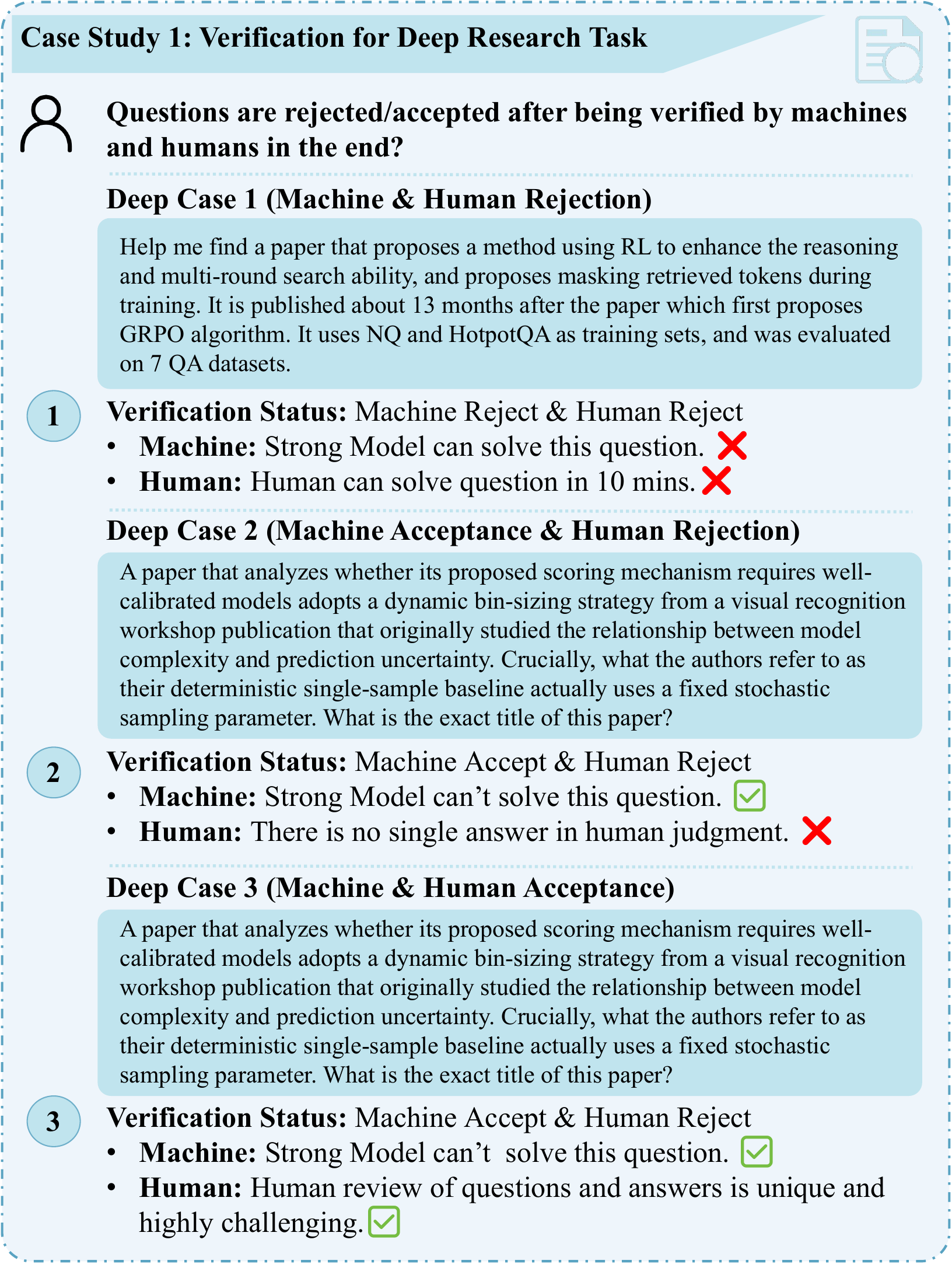}
    \caption{Representative rejection cases from the DeepSearch verification pipeline.}
    \label{fig:verification_case_deepsearch}
\end{figure}
To ensure that Deep Research instances are both non-trivial and well-posed, we adopt a four-stage adversarial verification pipeline: \textit{(i)} multi-query shortcut screening, \textit{(ii)} agent-based multi-step stress testing, \textit{(iii)} timed human retrieval, and \textit{(iv)} uniqueness auditing. An instance is retained only if it survives all four stages.
We provide some cases of verification, as shown in Figure~\ref{fig:verification_case_deepsearch}. 

\subsubsection*{Stage 1: Multi-Query Shortcut Screening}
For each candidate question, GPT-5.4~\cite{openai2026gpt54} rewrites the prompt into multiple search queries that vary clue granularity, lexical realization, and decomposition of the conjunctive constraints. These reformulations are issued to the same retrieval environment used at evaluation time. If the target paper can be directly surfaced or trivially identified from the returned results, the instance is discarded. This stage removes questions that appear obfuscated on the surface but still admit shallow lexical or paraphrastic retrieval.

\subsubsection*{Stage 2: Agent-Based Multi-Step Stress Testing}
We next stress-test the surviving instances with ReAct-style agents instantiated with Claude 4.6-Sonnet~\cite{anthropic2026sonnet46} and Gemini-3-Flash~\cite{deepmind2026gemini3flash}. Each agent is allowed to perform multi-turn search, browse candidate papers, and iteratively revise hypotheses in the same DeepXiv environment as the final evaluation. If either agent successfully solves the instance, the instance is removed. This stage filters out questions that resist one-shot retrieval yet remain too easy for contemporary agentic search.

\subsubsection*{Stage 3: Timed Human Adversarial Search}
For the remaining cases, human annotators conduct manual verification using the same search tools under a fixed 10-minute budget per instance. If a verifier can identify the answer within this time limit, the instance is rejected. This step prevents the benchmark from retaining questions that are only difficult for current models but still easily recoverable by a careful human searcher.

\subsubsection*{Stage 4: Uniqueness and Corpus-Level Audit}
Finally, annotators perform a corpus-level audit to verify that the conjunction of clues isolates exactly one valid paper. An instance is retained only if the gold answer is supported by explicit textual evidence and all plausible alternatives can be ruled out. If more than one paper satisfies the query, or if uniqueness cannot be established with confidence, the instance is discarded. Together, these steps ensure that Deep Research rewards genuine long-horizon evidence seeking rather than shortcut retrieval or annotation ambiguity.

\subsection{Wide Research Verification Pipeline}

\begin{figure}[ht]
    \centering
    \includegraphics[width=0.85\linewidth]{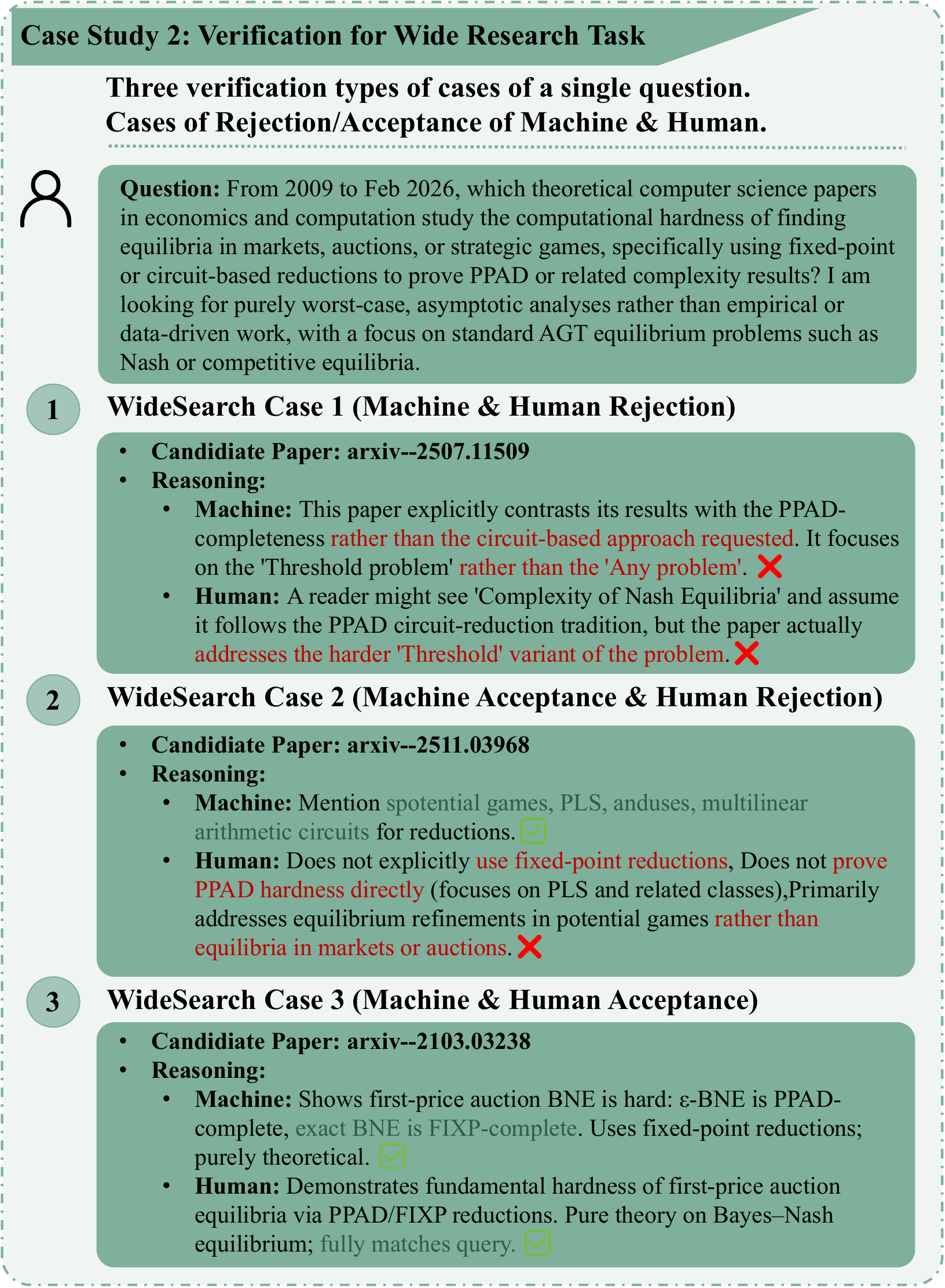}
    \caption{The Verification cases of WideSearch Task.}
    \label{fig:verification_case_widesearch}
\end{figure}

To ensure the factual grounding and condition-consistency of the Wide Research benchmark, we implement a rigorous three-stage verification pipeline: \textit{(i)} candidate expansion via multi-query search, \textit{(ii)} structural cleaning and temporal normalization, and \textit{(iii)} multi-model consensus verification with human-in-the-loop auditing. In this section, we give some cases of verification, as shown in Figure~\ref{fig:verification_case_widesearch}. 

\subsubsection*{Stage 1: Candidate Answer Supplement}
To mitigate false negatives and extend coverage, we systematically supplement candidate answers through large-scale arXiv search. For each question, an ensemble of LLMs(Gemini-3.1-pro~\cite{deepmind2026geminipro}, Claude 4.6-Sonnet~\cite{anthropic2026sonnet46}, GPT-5.4~\cite{openai2026gpt54}) generates approximately ten diverse search queries targeting methodological, domain-specific, and temporal facets. These queries are dispatched to the Jina search API, surfacing 31,734 candidate papers.

To manage downstream costs while preserving recall, we apply a lightweight LLM-based abstract screening to filter obvious off-topic results. Papers without available abstracts are retained conservatively. This process yields a 73.2\% retention rate (23,217 papers), with a per-question candidate mean of 33.0 ($\pm$15.2).

\subsubsection*{Stage 2: Data Cleaning and Temporal Normalization}
Candidate records undergo automated refinement to resolve structural inconsistencies:
\begin{itemize}
    \item \textbf{ArXiv ID Normalization:} Identifiers are parsed against canonical formats (e.g., \texttt{YYMM.NNNNN}). Spurious prefixes are stripped, while non-compliant IDs and their associated titles are purged to maintain field alignment.
    \item \textbf{Temporal Constraint Refinement:} For queries with an ambiguous ``2026'' upper bound, we invoke an LLM to calibrate the limit to month-level precision (e.g., ``before February 2026'') based on the latest ground-truth paper. This prevents incorrect penalization of agents retrieving papers published after the benchmark's effective cutoff.
\end{itemize}

\subsubsection*{Stage 3: Hybrid Machine-Human Verification}
The core verification relies on a content-grounded majority-voting scheme involving three independent frontier LLMs(Gemini-3.1-pro~\cite{deepmind2026geminipro}, Claude 4.6-Sonnet~\cite{anthropic2026sonnet46}, GPT-5.4~\cite{openai2026gpt54}). For each (question, paper) pair, the verifier retrieves the full text (truncated at 100,000 characters) from an indexed arXiv corpus. Models are explicitly instructed to be conservative, marking any condition as violated unless unambiguously demonstrated in the text. A final verdict is determined by a majority vote on the binary \texttt{satisfies} judgment.

To fortify the automated consensus, domain experts perform randomized audits by sampling 50\% of the papers that bypassed the multi-model filter. If the human-validated compliance rate for any query falls below a 75\% precision threshold, the entire candidate pool for that query is mandated to undergo a renewed cycle of multi-model voting and manual spot-checking. This iterative refinement persists until the threshold is met, neutralizing potential model hallucinations.

\paragraph{Verification Statistics.} Of the 23,217 screened candidates, 20,251 (87.2\%) were successfully retrieved for full-text analysis. Only 4,887 papers (24.1\%) passed the majority vote, reflecting the stringent multi-condition nature of the benchmark. This 75.9\% rejection rate confirms that WideSearch effectively distinguishes between general topical overlap and precise constraint satisfaction.

\begin{figure}[ht]
    \centering
    \includegraphics[width=\linewidth]{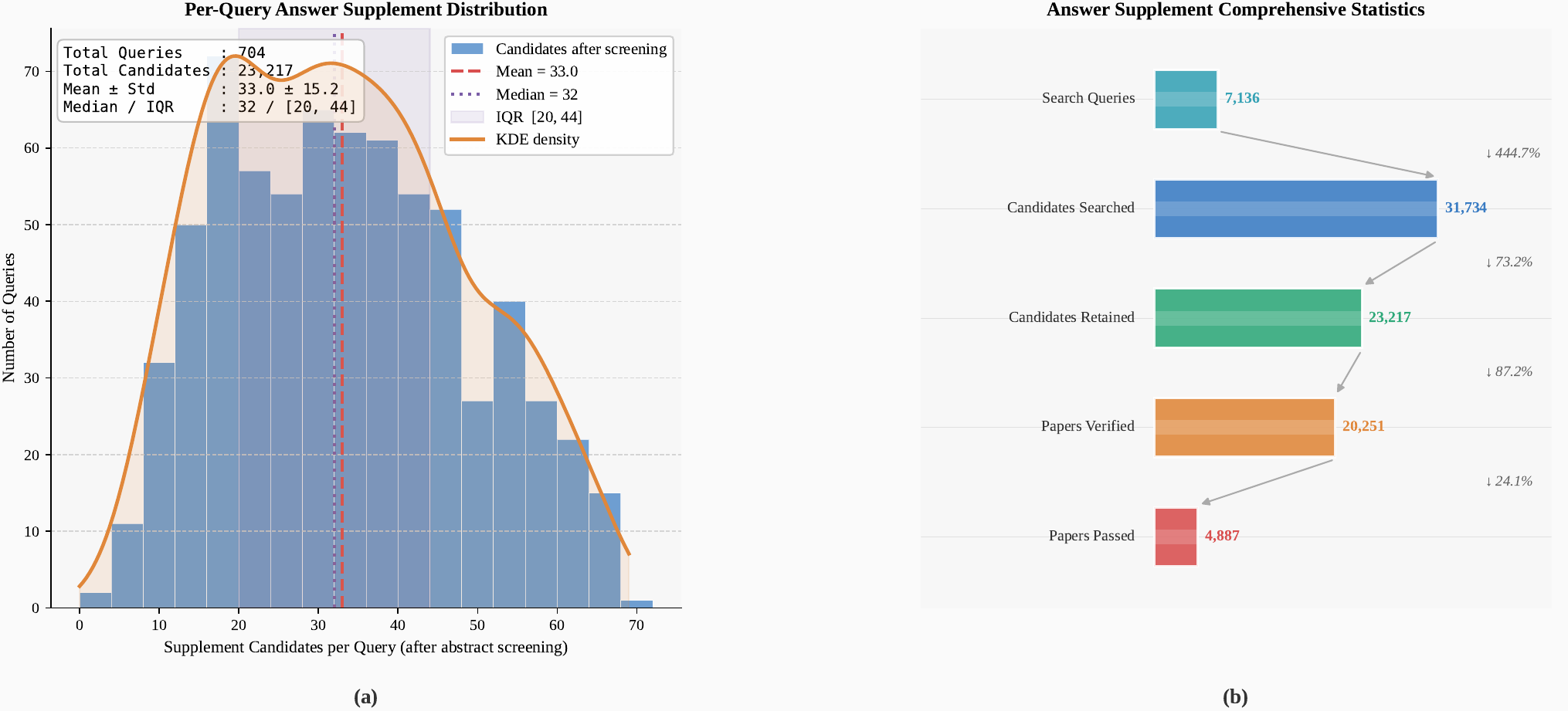}
    \caption{Statistics of the answer supplementation process. (a) Distribution of supplemented candidate papers per query. (b) Data pipeline waterfall showing cumulative reduction from LLM-generated queries through search, abstract filtering, multi-model verification, to final acceptance; percentages denote stage-wise retention rates.}
    \label{fig:answer_supplement_analysis}
\end{figure}

\section{Detailed Experiments}
\label{appendix:detailed experiments}
\subsection{Experimental Setup}

\subsubsection{Evaluation Framework}

\begin{figure}[ht]
    \centering
    \includegraphics[width=0.9\linewidth]{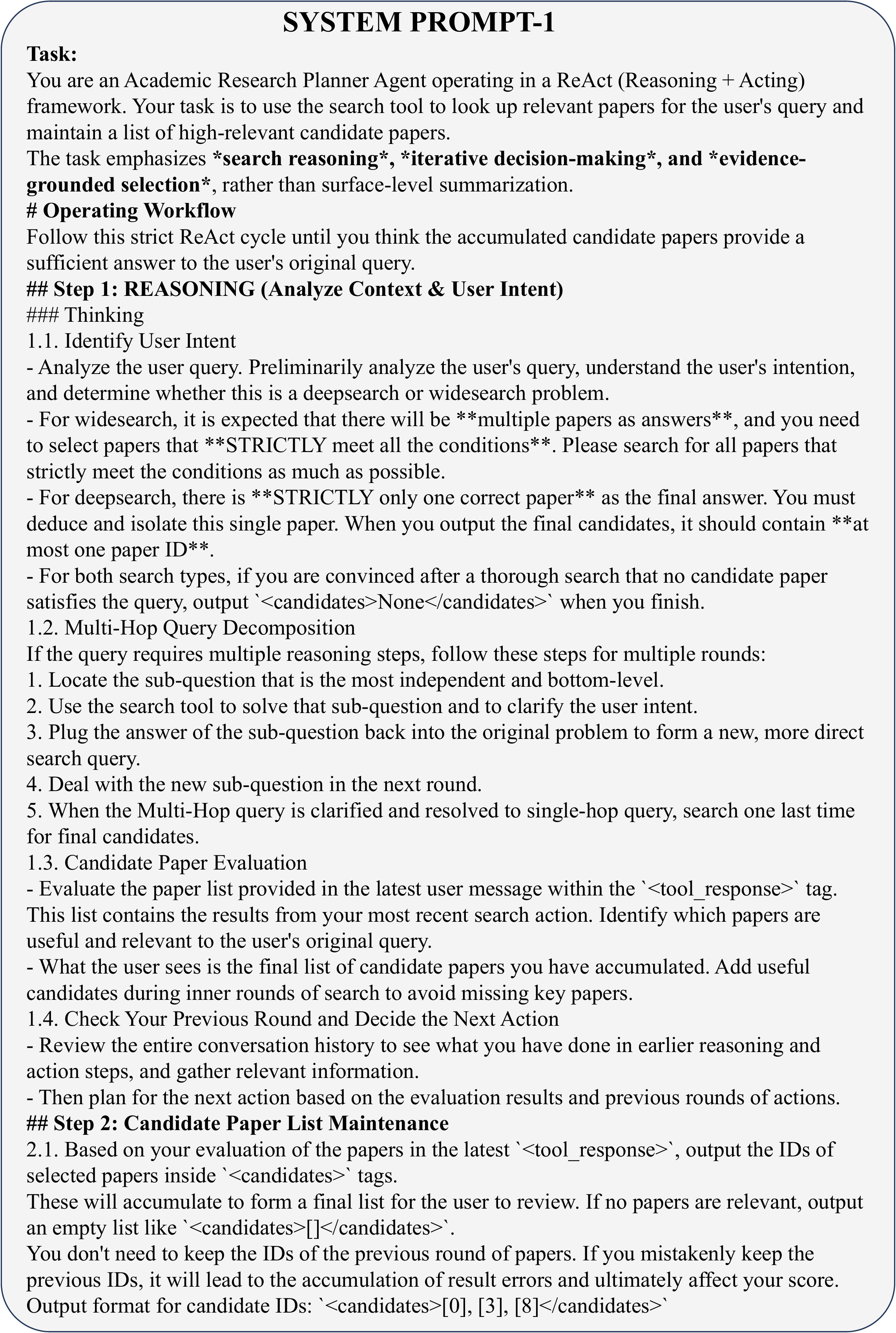}
    \caption{System prompt-1 of evaluation pipeline.}
    \label{fig:system-prompt-1}
\end{figure}

\begin{figure}[ht]
    \centering
    \includegraphics[width=0.9\linewidth]{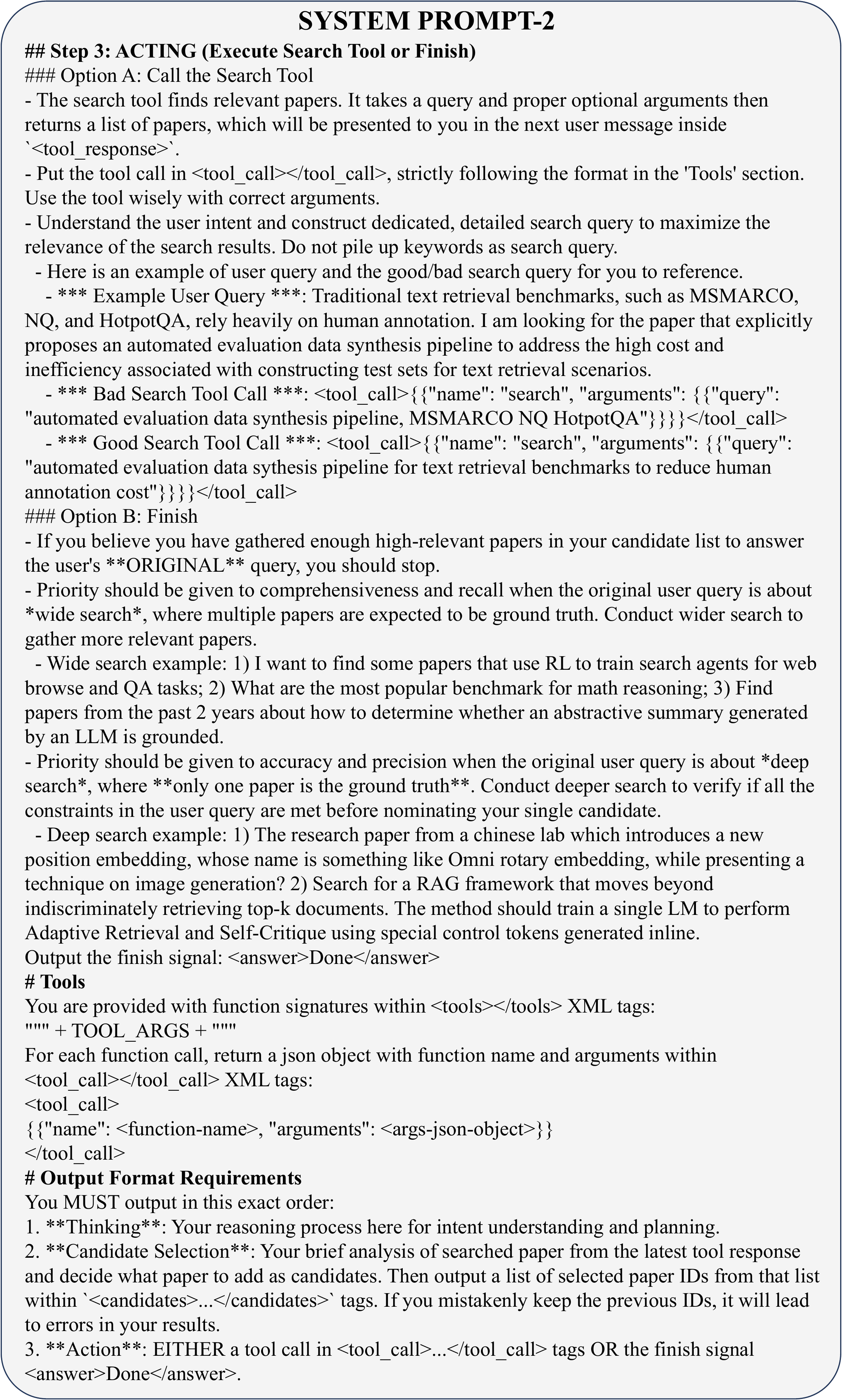}
    \caption{System prompt-2 of evaluation pipeline.}
    \label{fig:system-prompt-2}
\end{figure}

We evaluate agents in a single-tool ReAct loop: the model alternates between free-form reasoning, optional selection of paper IDs from the latest search, and either a structured search call or an explicit termination signal. The system prompt distinguishes \emph{wide} versus \emph{deep} research (multiple strict matches vs.\ at most one), encourages multi-hop decomposition when needed, and fixes the output order (thinking, \texttt{<candidates>}, then \texttt{<tool\_call>} or \texttt{<answer>Done</answer>}). System prompt is shown in Figure~\ref{fig:system-prompt-1} and Figure~\ref{fig:system-prompt-2}.

Each benchmark instance is processed asynchronously. Multiple independent trajectories per question are supported for pass@$k$ metrics; cross-instance concurrency is capped by a semaphore. Inputs and outputs are line-oriented JSON records; partial runs can be resumed by appending only missing questions. For each turn we log wall-clock duration, estimated input size, parsed action type, and the number of hits returned by the tool.

Termination occurs when the model emits the finish marker, when a soft context budget is exceeded (estimated with a tokenizer and a fixed threshold, after which a system warning forces final candidate selection without further tools), when the turn cap is reached (same warning pattern), or when the chat API fails (e.g., provider context limit or transport error). Parsed tool JSON errors are fed back as user-side error turns so the model can retry the format.

\subsubsection{Model and Runtime Hyperparameters}

Table~\ref{tab:infer-hparams} summarizes the main numerical settings. The planner uses a moderate sampling temperature and a long per-request HTTP timeout with a small number of automatic retries on failure. Completion length is bounded by a fixed cap on new tokens per turn unless overridden in code. Tool-side summarization uses the same completion wrapper with thinking disabled for deterministic JSON-style answers.

\begin{table}[t]
\centering
\caption{Default inference and tool-side hyperparameters (CLI flags override where noted).}
\label{tab:infer-hparams}
\small
\begin{tabular}{@{}l l@{}}
\toprule
\textbf{Quantity} & \textbf{Setting} \\
\midrule
Max agent turns & 30 (\texttt{--max-turns}) \\
Soft context budget (est.\ tokens) & $1.1\times 10^{5}$ \\
Papers shown from last search (UI to model) & 10 \\
Default search list size $k$ & 10 (per tool call) \\
Planner sampling temperature & 0.6 \\
Max new tokens per chat completion & 4096 \\
Chat request timeout & 1800\,s \\
Retries on chat failure & 5 \\
Trajectories per item (pass@$k$) & 1 (\texttt{-k}) \\
Concurrent benchmark items & 10 (\texttt{--max-workers}) \\
search HTTP timeout (paper / web) & 30\,s \\
Tool summarization temperature & 0.6 \\
\bottomrule
\end{tabular}
\end{table}

The planner may request extended reasoning via provider-specific chat extensions where enabled; tool summarizers disable that path to keep outputs short and schema-friendly.

\subsubsection{Tool Implementation}

\paragraph{Paper Search path.}
Search issues a POST with the natural-language query and $k$. Results are ranked hits with metadata and sectioned full text. For presentation to the agent, each hit is post-processed in parallel: author lists longer than ten names are truncated to five leading and five trailing names; a query-conditioned snippet is taken from the beginning of the stored body (first available section chunk); an auxiliary LLM then compresses that snippet into a query-focused ``evidence'' string that is shown as the paper's \texttt{search\_evidence}. Failures or timeouts yield an empty hit list for that call.

\paragraph{Web Search path.}
When this backend is selected, queries are prefixed to bias results toward arXiv. Two interchangeable connectors exist (programmatic web search vs.\ an alternate search+reader stack); the default path in the unified \texttt{search} interface uses the connector that returns both snippets and partial page text without a second fetch per URL. Each hit is capped to a large character window before the same query-conditioned summarization template is applied. The arXiv identifier shown to the agent is derived heuristically from the result URL when present.

\paragraph{Agent--tool interface.}
The exposed schema names the tool \texttt{search} (alias \texttt{PaperSearchTool} accepted at parse time). Arguments are a required \texttt{query} string and optional \texttt{top\_k}$=10$. The executor dispatches to the configured backend, reformats hits into a shared JSON structure, and injects them in the next user message inside \texttt{<tool\_response>} tags for the following turn.


\subsection{Detailed Results and Analysis}
\label{appendix:detailed_results}

\begin{table}[htbp]
\centering
\caption{Detailed performance metrics for the Deep Research task. Accuracy indicates the success rate of pinpointing the exact target paper. Token and Time metrics represent the average consumption per query.}
\label{tab:deepsearch_results}
\resizebox{\textwidth}{!}{%
\begin{tabular}{lccccc}
\toprule
\textbf{Model} & \textbf{Accuracy (\%)} & \textbf{Avg. Time (s)} & \textbf{Avg. Tokens} & \textbf{Avg. Turns} & \textbf{Avg. Tool Calls} \\
\midrule
Claude-Opus-4.6 & \textbf{9.39} & 642.4 & 19,112 & 28.1 & 27.1 \\
Gemini-3.1-Pro-Preview & 7.93  & 1221.4 & 12,206 & 24.4 & 22.8 \\
GPT-5.4 & 7.44 & 72.5 & 5,326 & 6.1 & 5.1 \\
Qwen3.5-397B-A17B & 6.97 & 3230.2 & 35,102 & 27.4 & 24.0 \\
Claude-Sonnet-4.6 & 6.96  & 588.4 & 21,540 & 27.5 & 26.5 \\
Seed-2.0-Pro & 6.80  & 310.1 & 14,444 & 22.9 & 21.8 \\
Kimi-k2.5 & 4.69  & 171.9 & 23,491 & 27.0 & 26.4 \\
DeepSeek-v3.2 & 4.21  & 405.7 & 21,575 & 28.8 & 25.8 \\
Qwen3.5-122B-A10B & 3.88  & 625.0 & 39,107 & 26.2 & 20.9 \\
Qwen3 Max & 3.24  & 166.0 & 8,771 & 6.1 & 5.1 \\
MiniMax M2.5 & 2.91  & 150.4 & 26,394 & 28.4 & 27.3 \\
Gemini 3 Flash & 2.75  & 236.9 & 14,052 & 15.9 & 14.9 \\
Qwen3.5-35B-A3B & 1.94  & 371.0 & 32,772 & 28.9 & 11.6 \\
\bottomrule
\end{tabular}%
}
\end{table}
\begin{table}[htbp]
\centering
\caption{Detailed performance metrics for the Wide Research task. The evaluation focuses on set-retrieval efficacy using average IoU, Recall, and Precision. Operational efficiency metrics (Time, Tokens, Turns) are also reported.}
\label{tab:widesearch_results}
\resizebox{\textwidth}{!}{%
\begin{tabular}{lccccc}
\toprule
\textbf{Model} & \textbf{Avg. IoU}  & \textbf{Avg. Time (s)} & \textbf{Avg. Tokens} & \textbf{Avg. Turns} & \textbf{Avg. Tool Calls} \\
\midrule
Gemini-3.1-Pro-Preview & \textbf{0.0931} & 235.33 & 15,045 & 4.55 & 3.49 \\
GPT-5.4 & 0.0812  & 115.98 & 12,905 & 3.69 & 2.68 \\
Seed-2.0-Pro & 0.0787  & 220.57 & 14,537 & 4.15 & 3.13 \\
DeepSeek-v3.2 & 0.0770  & 560.49 & 25,038 & 6.25 & 5.00 \\
Qwen3-Max & 0.0689 & 181.89 & 13,302 & 4.20 & 2.80 \\
Gemini-3-Flash & 0.0661  & 159.46 & 15,632 & 4.69 & 3.37 \\
Claude-Opus-4.6 & 0.0656  & 1032.90 & 82,165 & 26.44 & 25.44 \\
Kimi-k2.5 & 0.0623 & 353.92 & 30,017 & 8.35 & 7.31 \\
Claude-Sonnet-4.6 & 0.0496  & 1981.72 & 67,298 & 19.52 & 18.10 \\
Qwen3.5-397B & 0.0383 & 1027.32 & 26,586 & 7.11 & 5.91 \\
Qwen3.5-122B & 0.0276 & 594.87 & 19,411 & 5.39 & 3.97 \\
Qwen3.5-35B & 0.0271 & 274.66 & 50,911 & 12.83 & 9.71 \\
MiniMax-M2.5 & 0.0139 & 876.88 & 50,477 & 13.09 & 11.79 \\
\bottomrule
\end{tabular}%
}
\end{table}
This section presents the comprehensive quantitative outcomes of our empirical evaluation across all tested frontier models. To accurately reflect the distinct capability requirements of the two task paradigms, we report the performance metrics for Deep Research and Wide Research independently.

Table~\ref{tab:deepsearch_results} details the evaluation results for the Deep Research tasks. The primary metric is exact-match accuracy, which measures the agent's ability to successfully navigate multi-hop citation graphs and pinpoint the singular target document. Alongside accuracy, we report the average maximum turns, temporal cost, token consumption, and tool invocation frequency to characterize the efficiency and reasoning overhead of each model during deep exploration.

For Wide Research tasks, which essentially constitute a highly constrained set-retrieval problem, we expand the evaluation criteria to include Intersection over Union (IoU), Precision, and Recall. As demonstrated in Table~\ref{tab:widesearch_results} and Table~\ref{tab:widesearch_recall_precision}, breaking down the IoU into precision and recall provides critical insights into model behavior. For instance, while certain models exhibit relatively high recall, they concurrently suffer from extremely low precision, indicating a failure to strictly adhere to negative constraints and a tendency to return overly broad candidate sets. Unlike traditional top-$k$ or ranking-based search metrics, IoU evaluates the predicted set holistically. We specifically select IoU rather than the F1-score because the agent must effectively reason about, judge, and verify the coverage boundaries of scientific concepts. Furthermore, IoU imposes stricter penalties on false positives and false negatives, demonstrating greater sensitivity to boundary errors and small target sets compared to the F1-score. Consequently, this metric rigorously requires the agent to systematically enumerate all valid candidates while strictly avoiding out-of-scope inclusions, effectively measuring the capability of the agent in large-scale exploratory reasoning.

\begin{table}[htbp]
\centering
\caption{Average Recall and Average Precision decomposition for a representative selection of frontier models on the Wide Research task.}
\label{tab:widesearch_recall_precision}
\begin{tabular}{lcc}
\toprule
\textbf{Model} & \textbf{Avg. Recall} & \textbf{Avg. Precision} \\
\midrule
Claude 4.6 Sonnet & \textbf{0.3408} & 0.0586 \\
Kimi k2.5 & 0.2709 & 0.0839 \\
DeepSeek v3.2 & 0.1914 & 0.1297 \\
Gemini 3.1 Pro & 0.1781 & 0.1694 \\
GPT-5.4 & 0.1732 & 0.1346 \\
Seed 2.0 Pro & 0.1304 & \textbf{0.1787} \\
\bottomrule
\end{tabular}
\end{table}

A primary threat to validity in open-ended retrieval benchmarks is the potential incompleteness of the ground-truth set, which could unfairly penalize models that successfully retrieve valid, but unannotated, papers (i.e., false positives that are actually true). To rigorously assess the empirical validity of the low precision scores observed across all models, we conducted a manual auditing protocol to verify the nature of these ``extra'' retrieved answers. We randomly sampled the out-of-ground-truth papers submitted by three top-performing models and subjected them to blind human expert review against the strict query constraints. 

The audit revealed that 96\% of these extra predictions were indeed valid true negatives—meaning they objectively violated at least one explicitly stated constraint in the query (e.g., incorrect evaluation setting, mismatched methodology, or temporal violation). This extraordinarily high true negative rate substantiates the comprehensiveness of our dataset construction pipeline and confirms that the low precision scores are overwhelmingly attributable to model-side reasoning errors, constraint relaxation, or hallucination during tool usage, rather than missing gold-standard data. 

To systematically dissect these failure modes and provide a granular understanding of the operational bottlenecks within current LLM agents, the subsequent subsection will categorize the dominant error types and present concrete case studies of model reasoning trajectories.

\section{Detailed Error Analysis}
\label{appendix:detailed error analysis}
\begin{figure}[ht]
\centering
\includegraphics[width=0.9\linewidth]{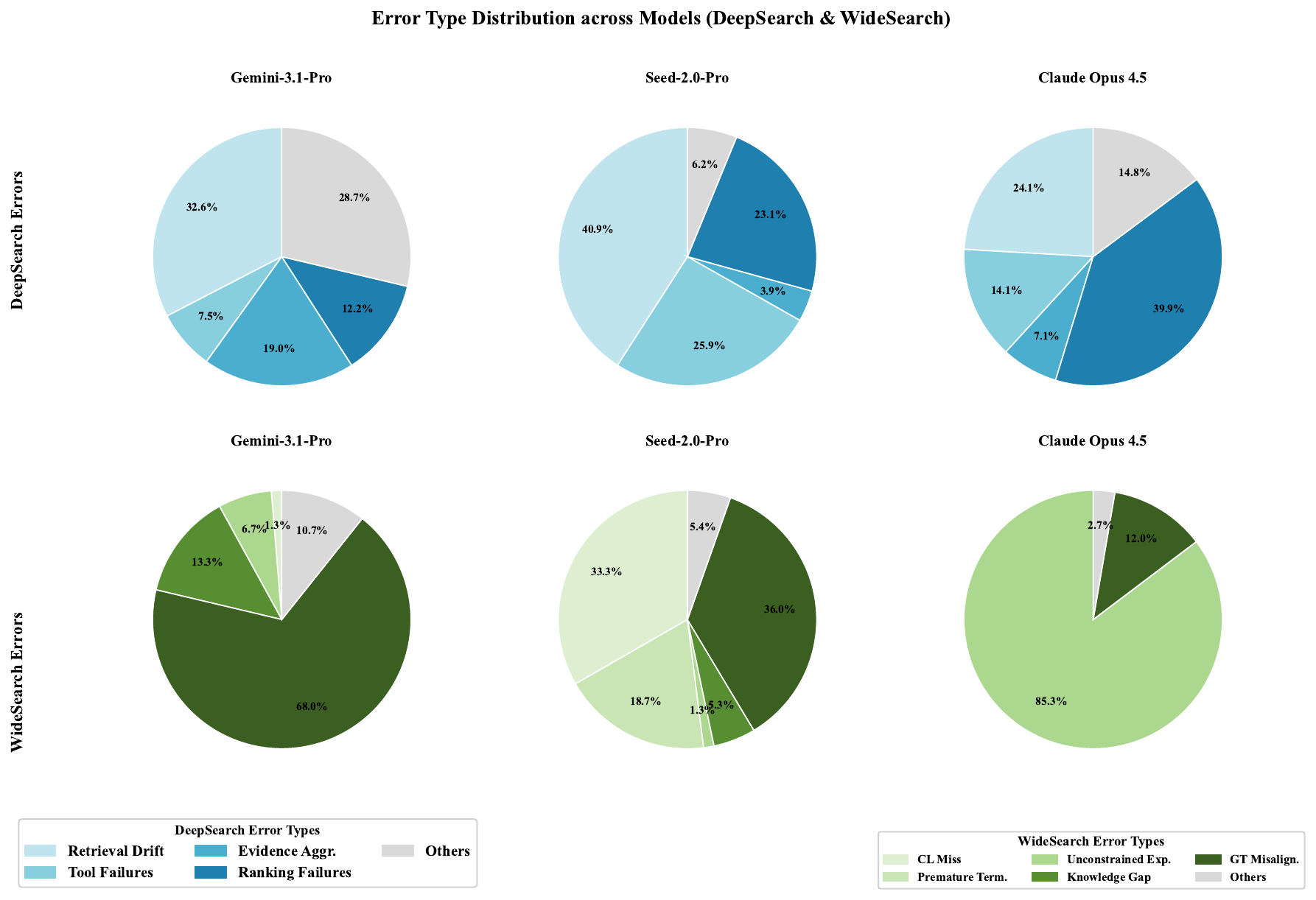}
\caption{Error-type distribution (as a percentage of manually labelled errors) for three agents on Deep Research (top row, blue palette) and Wide Research (bottom row, green palette). The gray \emph{Others} slice represents errors that do not match any defined category.}
\label{fig:error_pie}
\end{figure}

To diagnose systematic failure modes, we manually inspect a tratified sample of incorrect predictions from three frontier agents (Gemini-3.1-Pro, Seed-2.0-Pro, and Claude Opus 4.5) across both Deep Research and Wide Research tasks.  For each track we define a mutually exclusive taxonomy of error types Table~\ref{tab:error_types}), then measure the proportion of each type per model.  Figure~\ref{fig:error_pie} visualizes the distribution.
Because a small fraction of errors do not fit any category cleanly, the residual is grouped as \emph{Others}.
\begin{table}[h]
\centering
\small
\caption{Error type taxonomy for DeepSearch and WideSearch.}
\label{tab:error_types}
\renewcommand{\arraystretch}{1.3}
\begin{tabular}{p{0.2\linewidth} l p{0.7\linewidth}}
\toprule
\textbf{Error Type} & \textbf{Track} & \textbf{Definition} \\
\midrule
Retrieval Drift \& Semantic Confusion & Deep & The agent reaches the correct topical neighbourhood but commits to an adjacent paper that satisfies only a salient subset of the query constraints, e.g., confusing a cited anchor with the true target or conflating closely related sub-fields. \\

Tool Execution Failures & Deep & Malformed API calls, parser errors, or interrupted execution that consume search budget, disrupt evidence accumulation, and force the agent to fall back to unverified generic hypotheses. \\

Evidence Aggregation Failures & Deep & Sufficient evidence is retrieved but never consolidated: the agent fails to track which constraints are satisfied, which remain open, and which candidate best fits the full conjunction, resulting in redundant re-search and conservative non-commitment. \\

Candidate Ranking Failures & Deep & The gold paper is already in the candidate set, yet a different paper receives a higher rank. The bottleneck is reranking and consistency between intermediate reasoning and the final answer selection. \\

\midrule

Constraint Literalism-induced Miss & Wide & The agent applies query constraints too literally, discarding papers that satisfy the intent of the constraint but differ in surface form (e.g., year expressed as a range vs.\ an exact value). \\

Premature Search Termination & Wide & The agent stops exploring after finding a plausible candidate without exhaustively verifying whether a better-matching paper exists. \\

Precision-Unconstrained Candidate Expansion & Wide & The agent retrieves an overly broad candidate set without applying all specified constraints, inflating recall at the expense of precision and producing incorrect final answers. \\

Scientific Knowledge Coverage Gap & Wide & The model lacks domain knowledge required to interpret specialized terminology, limiting its ability to recognize the correct candidate even when it appears in the retrieved set. \\

GT Semantic Boundary Misalignment & Wide & The predicted answer is semantically close or factually adjacent to the ground truth but fails to match the exact entity the question intends. \\
\bottomrule
\end{tabular}
\end{table}

\subsection{Error Types}
\subsubsection{Deep Research Error Analysis}


\textbf{Retrieval drift and semantic confusion.}
The dominant failure mode is local semantic drift. Agents often reach the correct topical neighborhood but commit to a nearby paper that satisfies only a salient subset of the clues. Typical cases include confusing a cited anchor paper with the true target or conflating closely related subfields. The core issue is therefore not recall alone, but insufficient verification of the full constraint set.

\textbf{Tool-use and execution failures.}
A substantial fraction of errors arise from malformed tool calls, parser failures, or interrupted execution. These failures consume budget, break evidence accumulation, and prevent targeted follow-up on unresolved constraints. Once execution becomes unstable, agents often fall back to generic hypotheses rather than recover with disciplined verification, suggesting that robust tool use is part of the benchmarked capability itself.

\textbf{Evidence aggregation and termination failures.}
In this category, the agent retrieves many relevant clues but still fails to produce a sufficiently verified answer. All six trajectories end with an empty candidate list despite averaging 22.7 turns and 119.2 retrieved papers. The common problem is weak stateful reasoning over accumulated evidence: agents do not explicitly track which constraints have been satisfied, which remain unresolved, or which candidate best fits the conjunction, leading to redundant search and conservative non-commitment.

\textbf{Candidate selection and ranking failures.}
The final category occurs after successful retrieval: the gold paper is already in the candidate set, but the top-ranked prediction is incorrect. These cases indicate that, once candidate recall improves, exact reranking and consistency between intermediate reasoning and final selection become first-order bottlenecks.

\subsubsection{Wide Research Error Analysis}
\paragraph{GT semantic boundary misalignment.}
This constitutes the dominant failure mode for Gemini-3.1-Pro (68.0\%) and remains substantial for Seed-2.0-Pro (36.0\%). The predicted answer is semantically adjacent to the ground truth answer but does not match the exact entity intended by the question, which reflects ambiguity in the semantic scope of the query. The high prevalence of this error for Gemini-3.1-Pro suggests that the retrieval and ranking pipeline of the model is largely functional. The residual gap lies at the boundary between semantic plausibility and exact factual identity--a challenge that may require tighter grounding during the final selection step.

\paragraph{Precision-unconstrained candidate expansion.}
The error profile of Claude Opus 4.5 is dominated almost entirely by this failure type (85.3\%), a distribution unlike that of any other model in the study. The model retrieves a broad candidate set without systematically applying all specified constraints, thereby maximizing recall at the expense of precision. This phenomenon contrasts sharply with the ranking failure profile of the model in DeepSearch, suggesting that the same system over-selects under wide-coverage queries while struggling to rank candidates correctly under deep, multi-hop constraints.

\paragraph{Constraint literalism and premature termination.}
These two failure modes are conceptually linked: the agent applies surface-form constraints too strictly, rejects valid candidates early, and then terminates without broadening the search strategy. The joint occurrence of these errors implies that the search policy of Seed-2.0-Pro is inflexible at the constraint application stage and insufficiently persistent at the exploration stage.

\paragraph{Scientific knowledge coverage gap.}
Gemini-3.1-Pro accounts for the bulk of knowledge coverage gap errors (13.3\%). When the model cannot interpret domain-specific terminology, it fails to recognize the correct candidate even when the candidate is present in the retrieved set---an error type complementary to retrieval failure. This phenomenon points to limitations in domain-grounded understanding rather than in the search capability itself.

\subsection{Case Study}

\begin{figure}[ht]
\centering
\includegraphics[width=0.9\linewidth]{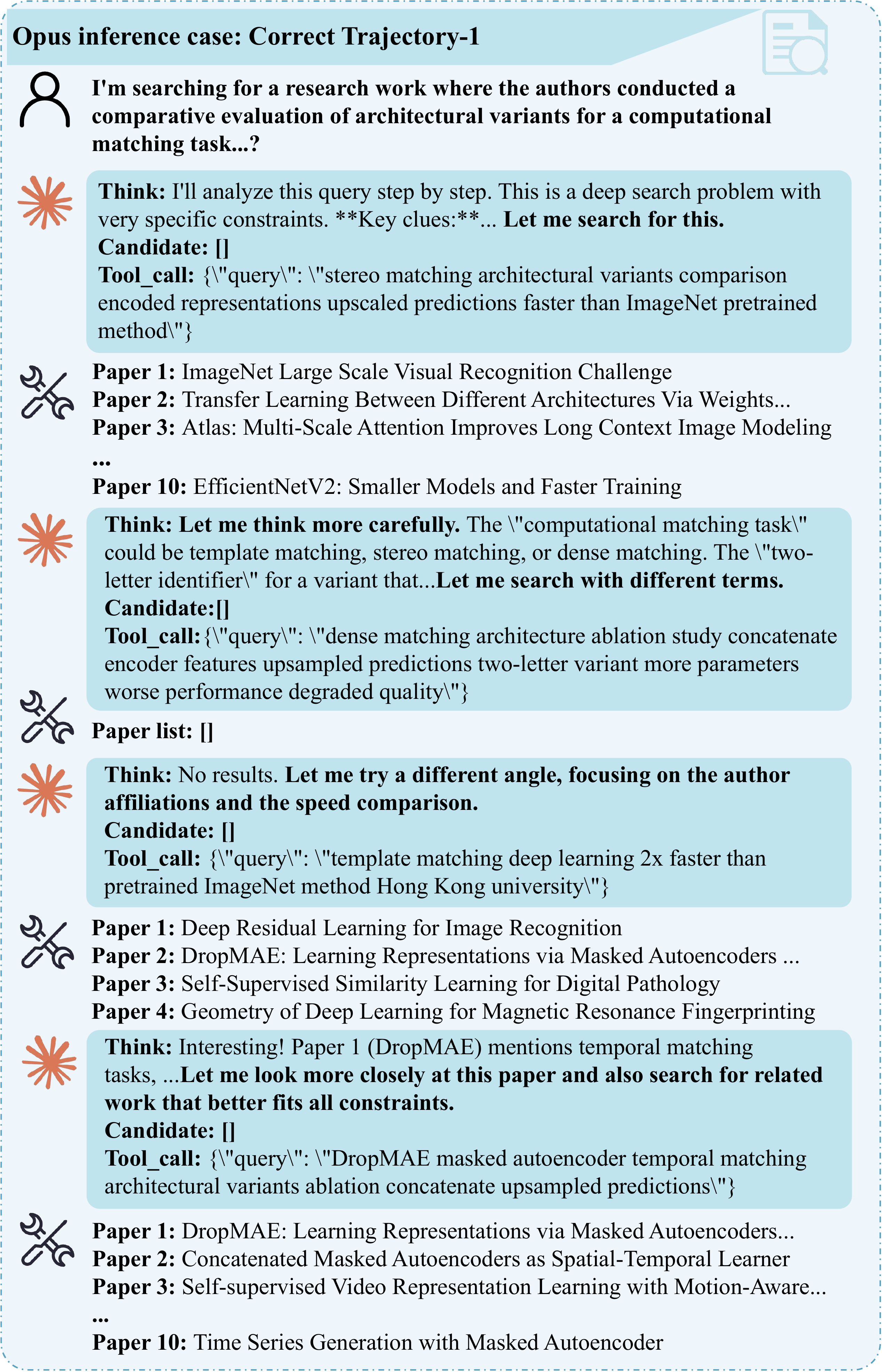}
\caption{The trajectory-1 of Opus in Deep Researh task. We omit detailed model responses and tool call results due to space constraints.}
\label{fig:deepsearch-cases-opus-correct-1}
\end{figure}

\begin{figure}[ht]
\centering
\includegraphics[width=0.9\linewidth]{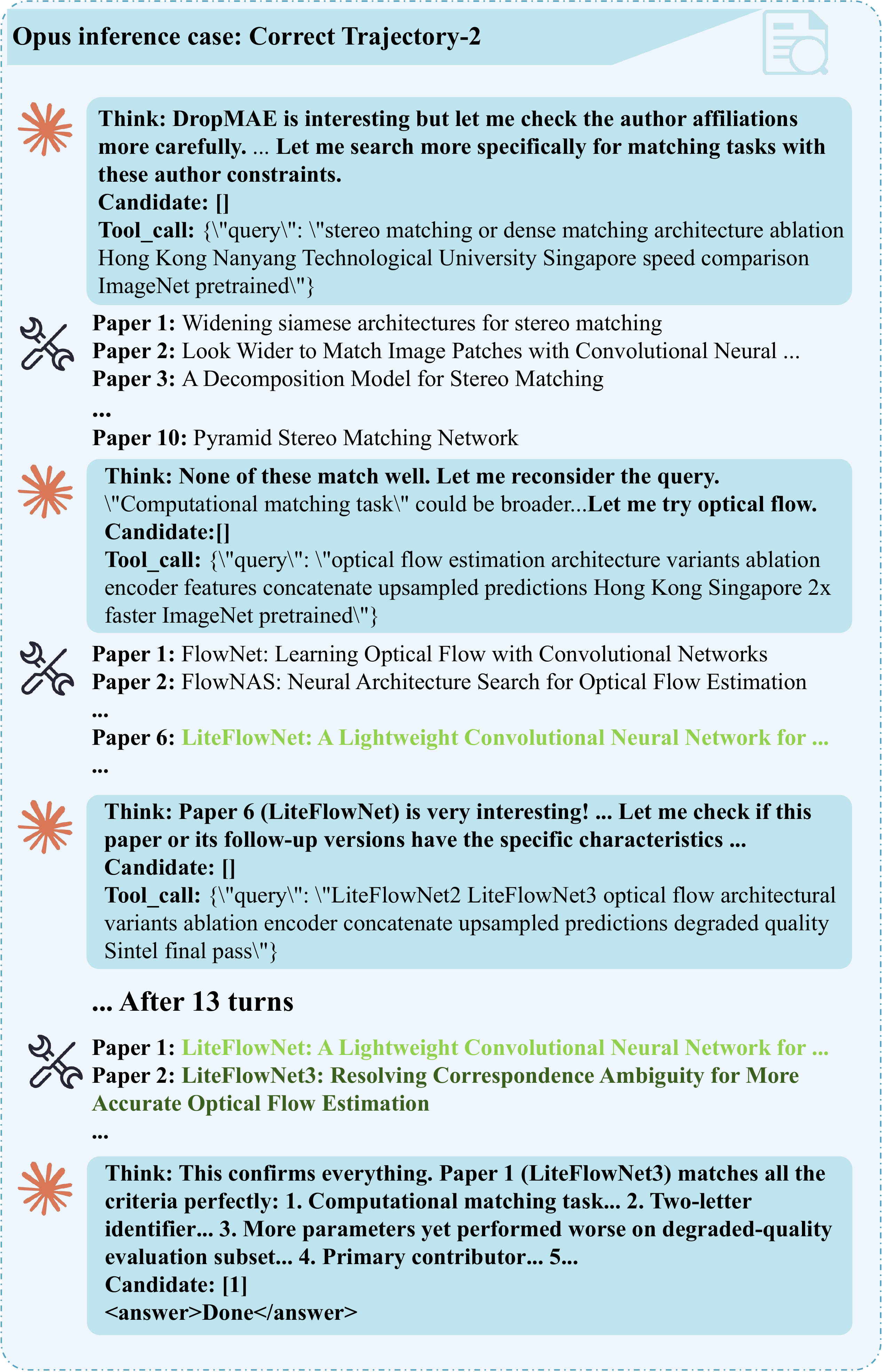}
\caption{The trajectory-2 of Opus in Deep Researh task. We omit detailed model responses and tool call results due to space constraints.}
\label{fig:deepsearch-cases-opus-correct-2}
\end{figure}

In this section, we present partial trajectories for both DeepSearch and WideSearch tasks. Figure~\ref{fig:deepsearch-cases-opus-correct-1} and Figure~\ref{fig:deepsearch-cases-opus-correct-2} illustrate a correct-solving trajectory of a model on the DeepSearch task, demonstrating its strong reasoning, self-reflection, and verification capabilities.

\section{Use of LLM}

We state the use of LLMs in this section. During dataset construction, we employ LLMs to assist with data creation, as detailed in Section~\ref{sec:academic_SAC}. For manuscript writing, we use LLMs for translation and linguistic refinement.

\end{document}